# Cross-Linguistic Transcription and Phonological Representation in the *Huìtóngguǎnxì Huáyíyìyǔ*

Ji-eun Kim[1*]

[1*]Department of Korean language and literature, Duksung Women's University, Seoul, South Korea.

Corresponding author(s). E-mail(s): smart173@duksung.ac.kr;

**Abstract**

**Purpose:** This study reconstructs the transcription principles of the *Huìtóngguǎnxì Huáyíyìyǔ* 會同館係 華夷譯語 (HHY), a set of multilingual wordlists compiled by the *Míng* 明 government between the fifteenth and sixteenth centuries for interpreter training. The study focuses on how Chinese characters were used to represent spoken forms of non-Chinese languages, and argues that HHY should be analyzed as a coherent multilingual transcription system rather than as a set of independent language-specific glossaries.
**Methods:** A substantial portion of HHY was digitized and aligned with Chinese phonological categories, or *yīnxì* 音系. Previous reconstructions of eight language sections - Korean, Japanese, Mongolian, Tibetan, Uyghur, Persian, Malay, and Cham - were critically reviewed and integrated into a unified comparative database. The analysis examines cross-linguistic correspondence patterns in *shēngmǔ* 聲母 and *yùnmǔ* 韻母 categories, with particular attention to the distinction between Main Transcription (MT) and Supplementary Transcription (ST).
**Results:** The results show that MT was governed primarily by the constraints of Chinese syllable structure. Sounds that could be accommodated within a Chinese syllable were normally represented by MT, whereas sounds that fell outside this structure satisfied the first condition for ST. ST, however, was not automatic. It was used only when such sounds also met additional phonetic conditions: they were voiced, voiceless continuants, or voiceless stops realized with audible release. The analysis further shows that Chinese phonological categories in HHY had transcriptional ranges that were broader and more flexible than their native Chinese values.
**Conclusion:** HHY therefore functioned as a systematic method of phonetic approximation, not as a direct projection of Chinese phonology onto foreign languages. Its cross-linguistic regularities provide comparative controls for interpreting individual language sections and offer a more precise framework for using HHY as evidence in Asian historical phonology.

**Keywords:** historical phonology, Chinese phonology, Asian languages, *Huìtóngguǎnxì Huáyíyìyǔ*

# 1 Introduction

Written records are indispensable for historical phonology, but they rarely provide direct access to speech (Lass 1997; Minkova 2015). Orthographic systems preserve phonological information only indirectly, because written forms must be interpreted through historically specific spelling practices and through the structure of the writing system itself (Lass 1997; Unger 2015). This problem is especially acute in the study of earlier Asian languages, where many languages are sparsely documented and where the available sources were not always designed as phonetic records. For this reason, transcriptional materials that explicitly attempted to record pronunciation deserve closer methodological attention. They can provide evidence that differs from both ordinary orthography and later comparative reconstruction.

This paper examines one such source: *Huìtóngguǎnxì Huáyíyìyǔ* 會同館係 華夷譯語 (HHY), a set of multilingual wordlists compiled by the *Míng* 明 government between the fifteenth and sixteenth centuries for the training of official interpreters (Lee 1957). Unlike literary translations, conventional lexical borrowings, or ordinary written records, HHY was intended to represent spoken forms of non-Chinese languages by means of Chinese characters (Lee 1957; Ishida 1931). It includes thirteen language sections, among them Korean, Japanese, Mongolian, Tibetan, Uygur, Persian, Malay, Cham, and several more poorly documented languages. HHY therefore occupies a distinctive position between orthography and phonetic notation: it is not an alphabetic transcription system, but neither is it an ordinary Chinese-character text.

The central claim of this paper is that HHY should be analyzed as a coherent Chinese character-based transcription system. The Chinese characters used in HHY did not simply reproduce their native Chinese phonological values. Rather, when they were used to transcribe foreign speech, their phonetic ranges were extended and reorganized according to the representational demands of the target languages. The transcriptional value of a given Chinese phonological category must therefore be reconstructed from its distribution across language sections, rather than inferred directly from Chinese phonology alone.

This point has broader methodological implications. Previous studies of HHY have generally focused on individual language sections, such as the Korean (Lee 1957, 1968; Kang 1995; Kwon 1995, 1998), Japanese (Watanabe 1961; Ōtomo 1968; Matsumoto and Ding 1997; Jiang 1998), Vietnamese (Chen 1966, 1967a,b, 1968a,b,c), Tibetan (Nishida 1963; Saita 1987), Uygur (Shōgaito 1982, 1984; Li 2019), Mongolian (Ochi 2004), Persian (Tasaka 1943a,b, 1944, 1951; Honda 1963), Malay (Edwards and Blagden 1931), and Cham (Edwards and Blagden 1939) sections. These studies have produced important reconstructions, but their findings remain scattered across different scholarly traditions and are not always easily accessible to an international readership. As a result, parallel transcriptional patterns across sections have rarely been treated as evidence for the structure of the HHY transcription system itself. Building on earlier language-specific work and on the comparative analysis of HHY transcription characters in Kim (2016), the present study shifts the focus from individual sections to the general principles governing the system as a whole.

The analysis is organized around two types of transcriptional elements: Main Transcription (MT) and Supplementary Transcription (ST). MT refers to the primary Chinese characters used to represent sounds that could be accommodated within the Chinese syllable structure of the HHY period. ST refers to additional Chinese characters used to represent sounds that could not be encoded by MT alone. The distinction is crucial because ST was not used randomly.

As this paper argues, ST was conditioned by the interaction of two factors: first, whether the target-language sound fell outside the representational capacity of Chinese syllable structure, and second, whether the sound was phonetically recoverable enough to be represented separately. In this sense, HHY transcription reflects not only Chinese phonological categories, but also a set of phonetic decisions about which non-Chinese sounds required additional notation.

To test this claim, the study digitizes and analyzes a substantial portion of HHY and aligns the transcription characters with Chinese phonological categories, or *yīnxì* 音系. Previous reconstructions of eight language sections - Korean, Japanese, Mongolian, Tibetan, Uygur, Persian, Malay, and Cham - are critically reviewed, converted where necessary into IPA-based representations, and integrated into a unified comparative database. The analysis then examines cross-linguistic correspondence patterns in Chinese *shēngmǔ* 聲母 and *yùnmǔ* 韻母 categories, with particular attention to the distribution of MT and ST.

The results show that HHY was governed by a consistent but flexible system of phonetic approximation. Sounds compatible with the Chinese syllable template were normally represented through MT. Sounds outside that template satisfied the structural condition for ST, but ST was used only when additional phonetic conditions were met: the relevant sound was voiced, a voiceless continuant, or a voiceless stop with audible release. These patterns show that HHY was neither a direct projection of Late *Míng* 明 Chinese phonology onto foreign languages nor an unsystematic set of ad hoc transcriptions. It was instead a constrained transcriptional system in which Chinese phonological categories were extended to encode non-Chinese speech.

By reconstructing these principles, this paper repositions HHY as a comparative resource for Asian historical phonology. The goal is not to reconstruct under-documented languages directly from HHY alone. Rather, the paper provides a framework for evaluating HHY evidence more systematically and for using well-documented language sections as controls in the interpretation of less secure sections. More broadly, the study shows how Chinese character-based transcription materials can be used not merely as isolated lexical sources, but as evidence for the phonological logic of premodern multilingual transcription.

# 2 Background

## 2.1 Four classes of *Huáyíyìyǔ*

*Huáyíyìyǔ* 華夷譯語 is a general term that indicates the dictionaries between the Chinese language and the languages of the neighboring regions, published by the Chinese government from the early *Míng* 明 dynasty and to the mid-*Qīng* 清 dynasty (Lee 1957; Li 2019) The dictionaries can be classified into four groups, as shown in Table 1.

No officially standardized terminology has yet been established for these four classes. In existing literature, however, they have been categorized in two main ways. First, following the chronological classification proposed by Ishida (1931), the four classes are labeled *kōshu* 甲種, *otsushu* 乙種, *heishu* 丙種, and *teishu* 丁種. Second, they are named according to their publishing institutions, except for the first class, whose publishers cannot be subsumed under a single designation: *zuìgǔběn* 最古本, *sìyíguǎnxì* 四夷館係, *huìtóngguǎnxì* 會同館係, and *huìtóngsìyíguǎnxì* 會同四譯館係 (Lee 1957; Kwon 1995; Kim 2016). As the second approach is more widely adopted in previous studies, this study follows that convention. In this study, *huìtóngguǎnxì* 會同館係, namely HHY is examined.

Table 1 Comparison of multilingual glossaries compiled in premodern China

| Publisher | Target language(s) | Transcription | Variety | Period |
|---|---|---|---|---|
| *huáyíyìyǔ*, *mǎshāyìhēi*, etc. | Mongolian | Chinese characters | Written language | 14th century |
| *sìyíguǎn* | Mongolian, Jurchen, Tibetan, Sanskrit, Persian, Tai, Uyghur, Burmese, Thai | Chinese characters + original script | Written language | 14th century |
| ***huìtóngguǎn*** | **Mongolian, Jurchen, Tibetan, Persian, Uyghur, Burmese, Thai, Korean, Ryukyuan, Japanese, Vietnamese, Cham, Shan** | **Chinese characters** | **Spoken language** | **15th–16th century** |
| *huìtóng sìyíguǎn* | (36 languages) | Chinese characters | Written language | 17th century |

We selected HHY because of its strong colloquial orientation, which allows for the analysis of phonetic and phonological properties of the target languages. As shown in Table 1, the other classes primarily document written languages. This difference reflects their distinct purposes of compilation. While HHY was compiled for the training of interpreters, the remaining classes were intended for the training of translators. The colloquial nature of HHY is particularly evident in materials on Mongolian, Uyghur, Persian, Tibetan, and Jurchen, whose spoken forms diverge substantially from their written traditions, which preserve much older stages of the languages (Lee 1957; Li 2019).

## 2.2 Composition of *Huìtóngguǎnxì Huáyíyìyǔ*

According to previous studies, six different manuscripts of HHY have been identified, currently preserved in various locations worldwide, including London, Japan (Seikadō Bunko 静嘉堂文庫; Nankaidō 阿波国; Inaba 稻葉君山), Taiwan, and Seoul (Seoul National University) (Lee 1957; Jiang 1998). Earlier studies also report the existence of two additional manuscripts formerly held in Hanoi and in Japan (Mito Akira 水戸彰), although these are no longer extant (Jiang 1998).

The remaining manuscripts do not contain identical sets of languages. Rather, each manuscript presents a different combination of target languages, as summarized in Table 2. The Nankaido manuscript appears to be the most comprehensive, containing 13 languages: Mongolian, Jurchen, Tibetan, Persian, Uyghur, Burmese, Thai, Korean, Ryukyuan, Japanese, Vietnamese, Cham, and Shan. For the Hanoi and Akira manuscripts, the language lists included in the table are limited to those that can be identified from previous studies; it remains unclear whether additional languages were originally included in these manuscripts.

In HHY, each language section constitutes a separate subsection containing several hundred lexical entries. These entries are further organized into thematic subsections, as illustrated in Table 4.

**Table 2** Coverage of target languages across manuscripts

| | London | Seikado | Nankaido | Inaba | Taiwan | SNU | Hanoi | Akira |
|---|---|---|---|---|---|---|---|---|
| **Korean** | Y | N | Y | Y | Y | Y | Y | Y |
| **Ryukyuan** | Y | N | Y | Y | Y | Y | | |
| **Japanese** | Y | Y | Y | Y | Y | Y | | |
| **Vietnamese** | Y | Y | Y | Y | Y | Y | | |
| **Cham** | Y | Y | Y | N | N | N | | |
| **Thai** | Y | Y | Y | Y | Y | Y | | |
| **Mongolian** | N | Y | Y | Y | Y | Y | Y | N |
| **Uyghur** | Y | Y | Y | Y | Y | Y | | |
| **Tibetan** | N | Y | Y | N | N | N | N | N |
| **Persian** | Y | Y | Y | N | N | N | | |
| **Malay** | Y | Y | Y | N | Y | Y | | |
| **Jurchen** | N | Y | Y | N | N | N | | |
| **Shan** | Y | Y | Y | Y | N | N | | |

Note: Y = yes (present); N = no (absent).

Among these, with regard to *shēng* 聲 in *shēngsèmén* (聲色門) of Cham, Edwards and Blagden (1939) suggested that it might instead be *yán* 顏. However, when the contents of this subsection (*mén* 門) are taken into account and compared with corresponding subsections in other languages that contain similar material, it can be confirmed that the correct reading is not (*yán*) 顏, but (*shēng*) 聲. This subsection includes colors-related vocabulary, such as *huáng* 黃 (C-539), *qīng* 青 (C-541), *hóng* 紅 (C-542), and *bái* 白 (C-543). Since Edwards and Blagden (1939, p. 87) themselves translated the name of this subsection (*mén* 門) as "Colours,"there is little doubt that this subsection introduces color-related vocabulary.

Although a set of core themes is shared across languages, certain themes appear only in specific language sections, and some languages exhibit unique thematic subsections absent from others. This variation likely reflects culture-specific lexical domains. For example, the subsections Diagram names and Sexagenary cycle occur exclusively in the Korean section, suggesting a particularly close cultural and intellectual connection between the Korean and Chinese communities at the time.

Meanwhile, each entry in HHY consists of either two or three rows, as illustrated in Table 3. The table shows the structure of entries from the Korean section (a) and the Tibetan section (b). Not all entries contain fully populated second or third rows, and some cells are left blank. The first row provides the semantic gloss, the second row transcribes the pronunciation of the native lexical form using Chinese characters, and the third row transcribes the pronunciation of the corresponding Sino-Xenic form. In the present study, the third row is excluded from analysis because Sino-Xenic vocabulary is heavily influenced by Chinese phonological systems and therefore differs fundamentally from native vocabulary in its phonological structure.

(a) **Table 3** Examples of HHY Entry Structure

| 天 | 日 | 月 | 星 | 風 |
|---|---|---|---|---|
| 哈嫩二 | 害 | 得一 | 別二 | 把論 |
| 忝 | 忍 | 臥 | 省 | 捧 |

(b)

| 天 | 地 | 日 | 月 | 風 |
|---|---|---|---|---|
| 難 | 薩 | 你麻 | 老瓦 | 弄 |

## 2.3 Chinese in *Huìtóngguǎnxì Huáyíyìyǔ*

It is important to determine the diachronic stage and dialectal background of the Chinese used in HHY. Since HHY does not provide clear information about when or where it was originally compiled, these questions cannot be answered from external records alone. Instead, the linguistic character of the Chinese reflected in HHY must be reconstructed through institutional context, internal linguistic evidence, and comparison with contemporary phonological sources.

The date of HHY can be approximated through two external reference points. The earliest possible date is 1408, when the *Huìtóngguǎn* 會同館, the institution responsible for compiling and publishing HHY, was officially established. Since HHY was produced as teaching material for interpreter training at the *Huìtóngguǎn*, its compilation must have begun after the institution had been established. The latest possible date is provided by the London manuscript of HHY, the most recent extant version, which was republished in 1549 (Ōtomo 1968). Together, these dates suggest that the initial compilation and publication of HHY took place sometime between 1408 and 1549.

Within this general timeframe, the dates of individual language sections can be further refined on the basis of linguistic features and relevant historical evidence. The Korean section appears to follow the general upper limit, but its lower limit can be raised, since the linguistic forms attested in the wordlist suggest a compilation date no later than the mid-fifteenth century (Lee 1957). The Japanese section, by contrast, follows the general lower limit but requires a later upper limit, because the Japanese bureau of the *huìtóngguǎn* 會同館 was not established until 1492, making an earlier compilation impossible (Ōtomo 1968). The Malay section likewise conforms to the general upper limit, but its lower limit can be narrowed to before 1511,

**Table 4** Theme-based subsections across the eight language sections

| Language | 1 | 2 | 3 | 4 | 5 | 6 | 7 |
|---|---|---|---|---|---|---|---|
| **Korean** | Astronomy | Geography | Time | Flowers and Trees | Birds and Animals | Buildings | Tools and Utensils |
| **Japanese** | Astronomy | Geography | Time | Flowers and Trees | Birds and Animals | Buildings | Tools and Utensils |
| **Mongolian** | Astronomy | Geography | Flowers and Trees | Birds and Animals | Buildings | Tools and Utensils | Humans |
| **Tibetan** | Astronomy | Geography | Toponymy | Time | Flowers and Trees | Birds and Animals | Buildings |
| **Uyghur** | Astronomy | Geography | Toponymy | Time | Flowers and Trees | Human matters | National events |
| **Persian** | Astronomy | Geography | Toponymy | Seasons | Flowers and Trees | Birds and Animals | Buildings |
| **Malay** | Astronomy | Geography | Time | Flowers and Trees | Birds and Animals | Buildings | Tools and Utensils |
| **Cham** | Astronomy | Geography | Time | Flowers and Trees | Birds and Animals | Buildings | Tools and Utensils |

| Language | 8 | 9 | 10 | 11 | 12 | 13 | 14 |
|---|---|---|---|---|---|---|---|
| **Korean** | Humans | Human matters | Body | Clothes | Colors | Jewels and Valuables | Food and Drink |
| **Japanese** | Humans | Human matters | Body | Clothes | Food and Drink | Jewels and Valuables | Literature and History |
| **Mongolian** | Human matters | Body | Clothes | Food and Drink | Jewels and Valuables | Literature and History | Colors |
| **Tibetan** | Tools and Utensils | Humans | Human matters | Body | Clothes | Food and Drink | Jewels and Valuables |
| **Uyghur** | Humans | Body | Clothes | Food and Drink | Tools and Utensils | Birds and Animals | Buildings |
| **Persian** | Tools and Utensils | Humans | Human matters | Body | Clothes | Food and Drink | Jewels and Valuables |
| **Malay** | Humans | Human matters | Body | Clothes | Food and Drink | Jewels and Valuables | Literature and History |
| **Cham** | Humans | Human matters | Body | Clothes | Food and Drink | Jewels and Valuables | Literature and History |

| Language | 15 | 16 | 17 | 18 | 19 | Subsections | Entries |
|---|---|---|---|---|---|---|---|
| **Korean** | Literature and History | Numerals | Sexagenary cycle | Diagram names | Miscellaneous | 19 | 596 |
| **Japanese** | Colors | Numerals | Directions | Miscellaneous | | 18 | 566 |
| **Mongolian** | Numerals | Miscellaneous | | | | 16 | 716 |
| **Tibetan** | Literature and History | Colors | Numerals | Miscellaneous | | 18 | 749 |
| **Uyghur** | Directions | Miscellaneous | Jewels and Valuables | Colors | Numerals | 19 | 839 |
| **Persian** | Literature and History | Colors | Numerals | Miscellaneous | | 18 | 674 |
| **Malay** | Colors | Numerals | Miscellaneous | | | 17 | 482 |
| **Cham** | Colors | Numerals | Miscellaneous | | | 17 | 601 |

Note: The numbered columns indicate the order of subsection titles within each language section.

the year in which the Kingdom of Malacca was conquered by Portugal; the linguistic and cultural background reflected in the Malay materials is consistent with the pre-conquest period (Edwards and Blagden 1931). In sum, all sections of HHY were compiled between 1408 and 1549, although the likely range differs somewhat depending on the language.

Having established when HHY was compiled, we can now turn to the dialectal background of the Chinese used in HHY. The Chinese reflected in most sections can be identified as *guānhuà* 官話, based on both institutional and phonological evidence. Institutionally, HHY was a government-sponsored project compiled by and for *Míng* 明 officials, in a context where *guānhuà* 官話 functioned as the standard spoken variety. Phonologically, the presence of *érhuà* 兒化 points to a northern variety of Chinese, since this feature is generally absent from southern dialects. The systematic loss of stop codas /-p/, /-t/, and /-k/ likewise supports a *guānhuà* 官話 affiliation, as this development is characteristic of northern Chinese varieties during the relevant period (Lee 2007).

An important exception to this general pattern is found in the Vietnamese section (Chen 1966). Unlike the other sections, it does not exhibit *érhuà* 兒化 and preserves stop codas /-p/, /-t/, and /-k/. These features allow a direct alignment between the Chinese transcriptions and corresponding Vietnamese stop codas, indicating that the Chinese used in this section reflects a southern variety rather than *guānhuà* 官話. This exception demonstrates that HHY did not apply a single Chinese dialect mechanically across all language sections, but rather adjusted its transcriptional practices to specific linguistic contexts.

## 3 Method

### 3.1 Reconstructing the Chinese Phonological System of *Huìtóngguǎnxì Huáyíyìyǔ*

As argued above, the Chinese used in HHY largely reflects Late *míng* 明 *guānhuà* 官話. A methodological difficulty arises, however, from the absence of phonological resources that directly reconstruct Late *Míng* 明 *guānhuà* 官話 as a unified system. The available materials instead span two adjacent periods in the history of Chinese, *zhōnggǔ* 中古 and *jìngǔ* 近古, necessitating a reconstruction strategy that can accommodate this transitional status. While there are differences among scholars, the history of Chinese is usually divided into four periods as Table 5:

**Table 5** Historical Stages of Chinese

| ***Shànggǔ***<br>Old | ***Zhōnggǔ***<br>Middle | ***Jìngǔ***<br>**Early Modern** | ***Xiàndài***<br>Modern |
|---|---|---|---|

3C 14C 20C

Although HHY belongs chronologically to the *jìngǔ* 近古 period, it was compiled relatively close to the late *zhōnggǔ* 中古 period. As a result, phonological sources associated with *zhōnggǔ* 中古 remain directly relevant to the present analysis. This is particularly significant because the *Zhōngyuányīnyùn* 中原音韻 (1324), one of the most comprehensive descriptions of spoken Chinese prior to the *Míng* 明 period, predates HHY by only several decades and continued to have influence during the Late *Míng* 明 period.

The literature referenced in this study can be divided into two groups. The first consists of historical phonological sources, including *zhōngyuányīnyùn* 中原音韻, *chóngdìng sīmǎwēngōng děngyùntújīng* 重訂司馬溫公等韻圖經, *sìshēngtōngkǎo* 四聲通考, *yùnlüèyìtōng* 韻略易通, and *xīrúěrmùzī* 西儒耳目資. The second comprises modern reconstructions based on these historical materials, such as Giles (1898); Minjungseorim (1966); Tōdō (1978); Kwon (1995); Li and Zhou (1999). Among these sources, this study relies primarily on *zhōngyuányīnyùn* 中原音韻 and Kwon (1995).

To reconstruct the Chinese phonological system underlying HHY, this study adopts a two-step alignment procedure. In the first step, all Chinese characters used in HHY were aligned with the phonological categories of *zhōngyuányīnyùn* 中原音韻, which was selected as the initial reference due to its extensive coverage of Chinese characters and its explicit orientation toward spoken forms. The *shēngmǔ* 聲母 and *yùnmǔ* 韻母 systems of *zhōngyuányīnyùn* 中原音韻 thus serve as the baseline for this initial alignment, with their phonetic values reconstructed in IPA based on previous studies (Minjungseorim 1966; Tōdō 1978; Kwon 1995; Li and Zhou 1999). In the second step, this preliminary alignment was systematically revised based on Kwon (1995), which reconstructs Late *Míng* 明 *guānhuà* 官話 on the basis of multiple historical sources and therefore more closely reflects the linguistic stage represented in HHY. Because the phonological systems documented in *zhōngyuányīnyùn* 中原音韻 underwent a series of changes between the fourteenth and seventeenth centuries, many of which were still in progress during the HHY period, these changes were treated not as categorical shifts but as gradual transitional tendencies in the alignment process.

More specifically, Table 6 is the *shēngmǔ* 聲母 system of *zhōngyuányīnyùn* 中原音韻. The phonetic value was reconstructed in IPA in the square brackets based on previous studies (Minjungseorim 1966; Tōdō 1978; Kwon 1995; Li and Zhou 1999).

**Table 6** The *shēngmǔ* 聲母 system of *zhōngyuányīnyùn* 中原音韻

| 幫 bang [p] | 滂 pang [ph] | 明 ming [m] | |
|---|---|---|---|
| 非 fei [f] | | | 微 wei [ɱ] |
| 端 duan [t] | 透 tou [th] | 泥 ni [n] | 來 lai [l] |
| 精 jing [ts] | 清 qing [tsh] | 心 xin [s] | |
| 章 zhang [tʃ] | 昌 chang [tʃh] | 山 shan [ʃ] | 日 ri [ʒ] |
| 見 jian [k] | 溪 xi [kh] | 疑 yi [ŋ] | 曉 xiao [x] |
| 云 yun [0] | | | |

The above system underwent five subsequent changes (Kwon 1995; Lee 2007). First, the *yí*-initial (*yímǔ* 疑母) was lost by the mid-fifteenth century. Second, the *wēi*-initial (*wēimǔ* 微母) merged into the zero-initial (*língshēngmǔ* 零聲母), realized as [w], by the seventeenth century. Third, the *zhěngchǐ*-series (*zhěngchǐyīn* 整齒音) became retroflexed after the sixteenth century. Fourth, the *jiàn*-series (*jiànxì* 見系) and the *jīng*-series (*jīngxì* 精系) were palatalized by the eighteenth century at the latest. With the exception of the loss of the *yí*-initial, these changes were not yet fully completed and were still in progress during the HHY period.

Next, Table 7 is the *yùnmŭ* 韻母 system of *zhōngyuányīnyùn* 中原音韻. The phonetic value was reconstructed in IPA in the square brackets based on previous studies (Minjungseorim 1966; Tōdō 1978; Kwon 1995; Li and Zhou 1999).

**Table 7** The yunmu system of *Zhōngyuán Yīnyùn*

| | 開口 kaikou | 齊齒 qichi | 合口 hekou | 撮口 cuokou |
|---|---|---|---|---|
| 東鐘 dongzhong | [uŋ] /uŋ/ | [iuŋ] /juŋ/ | | |
| 江陽 jiangyang | [aŋ] /aŋ/ | [iaŋ] /jaŋ/ | [uaŋ] /waŋ/ | |
| 支思 zhisi | [ɿ] /ï/ | [i] /jəj/ | | |
| 齊微 qiwei | [əi] /əj/ | | [uəi] /wəj/ | |
| 魚模 yumu | [u] /u/ | [iu] /ju/ | | |
| 皆來 jielai | [ai] /ai/ | [iai] /jaj/ | [uai] /waj/ | |
| 眞侵 zhenqin | [ən] /ən/ | [in] /jən/ | [uən] /wən/ | [iuən] /jwən/ |
| 寒山 hanshan | [an] /an/ | [ian] /jan/ | [uan] /wan/ | |
| 先天 xiantian | | [iɛn] /jen/ | | [iuɛn] /jwen/ |
| 桓歡 huanhuan | [ɔn] /on/ | | [uɔn] /won/ | |
| 蕭豪 xiaohao | [au] /aw/ | [iau] /jaw/ | | |
| 歌戈 gege | [ɔ] /o/ | [iɔ] /jo/ | [uɔ] /wo/ | |
| 家麻 jiama | [a] /a/ | [ia] /ja/ | [ua] /wa/ | |
| 車遮 chezhe | | [iɛ] /jə/ | | [iuɛ] /jwə/ |
| 庚青 gengqing | [əŋ] /əŋ/ | [iəŋ] /jəŋ/ | [uəŋ] /wəŋ/ | [iuəŋ] /jwəŋ/ |
| 尤侯 youhou | [əu] /əw/ | [iəu] /jəw/ | | |
| 侵尋 qinxun | [əm] /əm/ | [iəm] /jam/ | | |
| 監咸 jianxian | [am] /am/ | [iam] /jam/ | | |
| 廉纖 lianxian | | [iɛm] /jem/ | | |

The system in Table 7 underwent four subsequent changes (Kwon 1995; Lee 2007). First, the coda /-m/ merged into /-n/. This merger had already begun by the time of *zhōngyuányīnyùn* 中原音韻 and was completed before the sixteenth century. Second, the *huánhuān*-rime (桓歡韻) /on/ and the *xiāntiān*-rime (先天韻) /en/ lost their distinction during the sixteenth century. Third, the *érhuà*-rime (兒化韻) developed in northern dialects between the fifteenth and sixteenth centuries, as a result of which r-initial characters in the *zhīsī*-rime (支思韻), originally pronounced as [ʐɨ̧], began to be realized as [ər]. Fourth, the *gēgē*-rime (歌戈韻) /o/ and the *chēzhē*-rime (車遮韻) /e/ merged into [ɤ]. With the exception of the /-m/ to /-n/ merger, these changes were still in progress during the HHY period, although their effects are already observable in the HHY data (Kwon 1995).

## 3.2 Selecting reliable secondary resources

In this study, rather than reconstructing the HHY materials anew, we make extensive use of previous scholarship, that is, secondary resources. The reconstructions in the secondary resources are treated not merely as reference points, but as analytically substantive data that make it possible to examine HHY transcription practices on a broader, cross-linguistic scale.

At present, however, systematic phonological reconstructions are available for only eight of the thirteen language sections in HHY, even when older scholarship is taken into account. For this reason, the present analysis is limited to these eight languages: Korean, Japanese, Vietnamese, Tibetan, Uyghur, Mongolian, Persian, Malay, and Cham.

Core secondary references were selected according to three criteria. First, priority was given to studies that reconstructed an entire language section (i.e., a full language section), rather than focusing on individual or partial entries. Second, the study had to provide explicit phonetic reconstructions of the HHY entries, rather than relying solely on orthographic interpretation. Third, the reconstruction needed to demonstrate careful consideration of both the grammatical structure of the target language and the phonological constraints imposed by Chinese transcription practices. On the basis of these criteria, the core references listed in Table 8 were selected for each language section.

**Table 8** Previous reconstructions and their Chinese phonological references

| **Language** | **Reconstruction** | **Manuscript** | **Chinese Phonology** |
|---|---|---|---|
| Korean | Kwon (1995) | Inaba | *Zhōngyuányīnyùn*<br>*Wēngōng Děngyùn Tújīng*<br>*Sìshēngtōngkǎo*<br>*Yùnlüèyìtōng* |
| Japanese | Matsumoto and Ding (1997) | Nankaido<br>London<br>Inaba<br>Seikado | *XīrúĚrmùzī* |
| Vietnamese | Chen (1966) | Nankaido | *Zhōngyuányīnyùn*; *Yùnlüèyìtōng*<br>*Zhōnghuáxīnyùn* |
| Tibetan | Nishida (1963)<br>Saita (1987) | Seikado | *Wēngōng Děngyùn Tújīng* |
| Uyghur | Shōgaito (1984) | Nankaido<br>Seikado<br>London | *Wēngōng Děngyùn Tújīng* |
| Mongolian | Ochi (2004) | Seikado<br>Nankaido<br>Inaba | *Wēngōng Děngyùn Tújīng* |
| Persian | Honda (1963) | London | *Wēngōng Děngyùn Tújīng* |
| Malay | Edwards and Blagden (1931) | London | Giles (1898) |
| Cham | Edwards and Blagden (1939) | London | Giles (1898) |

Although each of the selected studies in Table 8 provides reconstructions of the phonetic values represented by Chinese characters in HHY, they rely on different reference systems for Chinese phonology. Among these, *sīmǎ wēngōng děngyùn tújīng* 重訂司馬溫公等韻圖經 (1606), which is cited most frequently across the selected studies, is a traditional Chinese phonological chart (*yùntú* 韻圖, reflecting northern *guānhuà* 官話 pronunciation of the early seventeenth century. As noted in Kwon (1995), this work serves as a major reference for reconstructing the northern phonological system of the fifteenth and sixteenth centuries, second only to *zhōngyuányīnyùn* 中原音韻. Meanwhile, *yùnlüèyìtōng* 韻略易通 (1442), a fifteenth-century source, preserves checked syllable codas (-p, -t, -k) and maintains a split within the *yúmú*-rime (*yúmúyùn* 魚模韻), features that distinguish it from *zhōngyuányīnyùn* 中原音韻. While these characteristics limit its applicability to certain aspects of the phonological system, the work was nonetheless consulted as a supplementary reference for issues other than initial consonants. In the case of the Vietnamese section, whose *yùnmǔ* 韻母 reflects southern phonological characteristics distinct from those of other language sections, such southern-oriented rime books could function as primary reference sources.

### A. Japanese section *Rìběnguǎnyìyǔ* 日本館譯語

Research on the Japanese section, *rìběnguǎnyìyǔ* 日本館譯語 has advanced considerably in Japanese linguistics, and Watanabe (1961) already provided a reliable reconstruction of the entire lexical inventory at an early stage of the field. Among the various sections of the *huáyíyìyǔ* 華夷譯語 examined in this study, the Japanese section therefore constitutes one of the most dependable points of comparison.

The principal reference adopted here is Matsumoto and Ding (1997). Building on earlier reconstructions, Matsumoto and Ding (1997) differ from previous studies in that they provide phonetic reconstructions using the International Phonetic Alphabet (IPA). This is especially relevant because HHY, unlike other classes of *huáyíyìyǔ* 華夷譯語, records spoken forms rather than written norms. It should therefore be approached from a phonetic perspective. In this respect, Matsumoto and Ding (1997) offer a particularly useful basis for comparison.

This does not mean, however, that the analysis can rely exclusively on Matsumoto and Ding (1997). Scholarly views do not agree on every entry in the Japanese section, and several important transcription characters remain disputed. Accordingly, this study takes Matsumoto and Ding (1997) as its primary point of reference, while also consulting *kana*-based reconstructions and other previous studies where necessary (Ōtomo 1968; Honda 1963; Jiang 1998). In addition, relevant research on Middle Japanese is used to further evaluate and cross-check the proposed interpretations.

### B. Vietnamese section *Ānánguǎnyìyǔ* 安南館譯語

Comprehensive research on the Vietnamese section, *ānnánguǎnyìyǔ* 安南館譯語, was conducted primarily by Chen Ching-ho in a series of studies (Chen 1966, 1967a,b, 1968a,b,c). These studies examine the Vietnamese section from multiple perspectives, ranging from bibliographical issues to the phonological system reflected in the transcribed Vietnamese forms. Because they provide a detailed and comprehensive treatment of the material, they remain the most authoritative studies on the Vietnamese section to date, and their major conclusions have not been seriously challenged in subsequent scholarship.

One limitation, however, is that the reconstruction of entries in the Vietnamese section has not been extensively reexamined by a broad range of scholars. To compensate for this relative lack of independent verification, the present study additionally consults previous research on the history of Vietnamese phonology.

Furthermore, the Vietnamese section is used in the present study only as a comparative source for investigating the phonetic values of fifteenth-century Korean initial consonants reflected in the Korean section. According to Chen (1967a), the Vietnamese section differs from several other sections, including the Korean section, in that its use of codas shows a particularly strong influence from southern varieties of Chinese phonology.

#### C. Tibetan section *Xīfānguǎnyìyǔ* 西番館譯語

Reconstruction of the entries in the Tibetan section, *xīfānguǎnyìyǔ* 西番館譯語, has primarily been carried out by two researchers. Nishida (1963) provided the first comprehensive study of the Tibetan section. In addition to bibliographical analysis, Nishida (1963) reconstructed the literary and colloquial Tibetan forms corresponding to individual entries on the basis of modern Tibetan dialects, written Tibetan, and separate Tibetan glossaries. Through this approach, he clarified both the phonological and lexical characteristics of the Tibetan reflected in the Tibetan section.

One limitation of Nishida (1963), however, is that the study relied exclusively on the Awa manuscript tradition. Subsequently, Saita (1987) largely followed Nishida (1963)'s reconstruction of Tibetan forms while revising portions of the analysis on the basis of the Seikadō manuscript. The present study therefore adopts Nishida (1963) as its principal reference while incorporating revisions proposed in Saita (1987) when they are supported by comparison between different manuscripts.

Although the Tibetan section has thus been reconstructed by two scholars, its interpretations cannot yet be regarded as sufficiently verified. To compensate for this limitation, the present study additionally consults a broader range of scholarship on Tibetan historical phonology.

#### D. Uyghur section *Wèiwúérguǎnyìyǔ* 畏兀兒館譯語

Comprehensive research on the Uyghur section, *wèiwúérguǎnyìyǔ* 畏兀兒館譯語, has primarily been conducted by a single Japanese scholar. Reconstruction of the complete lexical inventory was carried out only in Shōgaito (1984). In an earlier study, Shōgaito (1982) had already discussed the linguistic character of the Turkic language reflected in the Uyghur section, although without reconstructing the complete lexical inventory.

In Shōgaito (1984), the colloquial Uyghur forms corresponding to individual entries were reconstructed on the basis of modern Uyghur dialects, Mongolic languages, written Mongolian, and other Chinese transcription materials of Uyghur. Through this comparative approach, the study provided a detailed account of the phonological characteristics of the Uyghur reflected in the Uyghur section. The major conclusions of Shōgaito (1984) have not been seriously challenged in subsequent scholarship.

One limitation, however, is that the reconstructed entries in the Uyghur section have not undergone extensive verification by a broad range of scholars. To compensate for this limitation, the present study additionally consults broader research on Uyghur and Mongolic historical phonology.

#### E. Mongolian section *Dádánguǎnyìyǔ* 韃靼館譯語

Comprehensive research on the Mongolian section, *dádánguǎnyìyǔ* 韃靼館譯語, has primarily been conducted by Ochi (2004). In addition to bibliographical issues such as textual format and relationships among manuscript traditions, Ochi (2004) also examined the phonological characteristics of the Mongolian reflected in the text in considerable detail. Rather than relying exclusively on reconstructed Chinese readings, Ochi (2004) approached the material with full consideration of the character of the HHY as a transcription of spoken language. Through detailed comparison with written Mongolian and modern Mongolian dialects, Ochi (2004) clarified important aspects of the phonological system reflected in the Mongolian section.

One limitation, however, is that the lexical reconstructions proposed for the Mongolian section have not undergone extensive verification by a broad range of scholars. The present study therefore additionally consults broader research on Mongolian historical phonology.

At the same time, the Mongolian section is used in the present study only as a comparative source for investigating the transcription pattern of onsets and codas in HHY. This is because, unlike several other major studies discussed above, Ochi (2004) did not provide reconstructions for the complete lexical inventory, but instead systematically identified the Mongolian sounds represented by individual Chinese transcription characters. Also, if the reconstruction of the transcriptional system in Ochi (2004) had been based on assumptions identical to those adopted in the present study, the analysis could have been applied more directly. However, Ochi (2004)'s reconstruction of the phonological values of the transcription characters relied primarily on the *chóngdìng sīmǎ wēngōng děngyùn tújīng* 重訂司馬溫公等韻圖經, which reflects seventeenth-century Beijing Mandarin. It is therefore possible that the reconstructed values of the transcription characters differ in some respects from those proposed in the present study.

Relatively simple initials can still be reinterpreted on the basis of Ochi (2004)'s examples within the analytical framework adopted here. Codas, however, are considerably more difficult to analyze in this way because of their greater structural complexity. Moreover, since the study did not systematically examine the entire lexical inventory with respect to initials and codas, conclusions drawn from comparison with the Mongolian section should be treated with particular caution.

**F. Persian section *Huíhuíguǎnyìyǔ* 回回館譯語**

Comprehensive research on the Persian section, *huíhuíguǎnyìyǔ* 回回館譯語, has primarily been conducted by two researchers. Substantial work on the Persian section had already begun in a series of studies by Tasaka Kōdō (Tasaka 1943a,b, 1944, 1951). These studies, however, reconstructed only forty-nine entries from the astronomy section. Honda (1963) later overcame this limitation by examining both the *Huìtóngguǎnxì* 會同館係 and *Sìyíguǎnxì* 四夷館系 Persian sections and reconstructing the complete lexical inventory and example phrases in Persian. The present study refers only to Honda (1963)'s reconstruction of the Persian section in HHY.

Although the Persian section has thus been reconstructed by two scholars, its interpretations cannot yet be regarded as sufficiently verified. Moreover, unlike previous studies on several other sections, Honda (1963) reconstructed the entries without providing a systematic analysis of the transcriptional system itself. Another difficulty is that the study presents phonemic rather than phonetic reconstructions, meaning that interpretation of the actual phonetic values requires additional knowledge of historical Persian phonology.

Despite these limitations, the Persian section remains important for the present study. Persian permits word-final consonant clusters and therefore provides important evidence for analyzing the phonetic value and transcribing rules of coda consonants. To compensate for the limitations of previous research, the present study additionally consults a broad range of scholarship on Persian historical phonology.

#### G. Malay and Cham sections *Mǎnlàjiāguǎnyìyǔ* 滿剌加館譯語 and *Zhānchéngguǎnyìyǔ* 占城館譯語

The reconstructions of the Malay section, *mǎnlàjiāguǎnyìyǔ* 滿剌加館譯語, and the Cham section, *zhānchéngguǎnyìyǔ* 占城館譯語, were carried out by the same researchers (Edwards and Blagden 1931, 1939). Although a substantial number of transcription characters remained unresolved in these studies, they nevertheless remain the only works to attempt reconstruction of the complete lexical inventories of the Malay and Cham sections.

At the same time, Edwards and Blagden (1931) and Edwards and Blagden (1939) are relatively old studies, and many of their reconstructions appear to have been based primarily on modern Malay and Cham without sufficiently detailed consideration of historical Chinese phonology. As a result, conclusions derived solely from comparison with the Malay and Cham sections cannot always be accepted without caution.

The present study therefore supplements these earlier reconstructions through extensive consultation of additional scholarship on Malay and Cham, especially diachronic studies. Through this approach, the study seeks to compensate as far as possible for the limitations of the existing reconstructions.

#### H. Korean section *Cháoxiǎnguǎnyìyǔ* 朝鮮館譯語

Research on the Korean section, *cháoxiǎnguǎnyìyǔ* 朝鮮館譯語, began with its introduction by Ogura (1941), although the section did not become widely recognized as an important source for Korean historical linguistics until Lee (1957) established the chronological upper and lower bounds of its compilation. Subsequent research on the Korean section can generally be divided into comprehensive studies and studies focusing on specific linguistic issues. The present study discusses only the former, referring to more specialized studies only where directly relevant.

The first comprehensive study of the Korean section was conducted by Lee (1968). In addition to examining bibliographical issues such as manuscript relationships and the dating of the text, Lee (1968) investigated linguistic features reflected in the Korean section, including aspects of fifteenth-century Korean phonology and vocabulary. One limitation of the study, however, is that the complete lexical inventory was not fully reconstructed. In addition, the reconstruction of the second and third rows relied primarily on *měnggǔzìyùn* 蒙古字韻, a rime book compiled in 1308, without reconstructing the phonological system of fifteenth-century Chinese itself.

Kang (1995) improved upon this limitation by consulting *yùnlüèyìtōng* 韻略易通 (1442) and *yùnlüèhuìtōng* 韻略匯通 (1642), phonological sources chronologically closer to the fifteenth-century northern Chinese reflected in the Korean section. In addition to collating multiple manuscripts, Kang (1995) reconstructed and glossed the complete lexical inventory and systematically examined a wide range of phonological and grammatical features of fifteenth-century Korean. Nevertheless, the Chinese phonological sources employed in the study were still insufficient for a precise reconstruction of the fifteenth-century northern Chinese phonological system reflected in HHY.

These limitations were largely overcome by Kwon (1995, 1998). Using *zhōngyuányīnyùn* 中原音韻, *chóngdìng sīmǎ wēngōng děngyùn tújīng* 重訂司馬溫公等韻圖經, and the colloquial readings of *sìshēngtōngkǎo* 四聲通考 as primary references, supplemented by *yùnlüèyìtōng* 韻略易通 and *yùnlüèhuìtōng* 韻略匯通, Kwon (1995, 1998) reconstructed the fifteenth-century northern Chinese phonological system through a phonological approach and used it as the basis for interpreting the second and third rows of the Korean section. On this basis, the study also reconstructed important aspects of the fifteenth-century Korean phonological system itself. With the exception of a small number of unresolved items, Kwon (1995, 1998) largely completed the comprehensive reconstruction of the Korean section, and no comparably comprehensive study has appeared since.

It should be noted, however, that the reconstruction of the Korean section differs in character from many of the reconstructions available for other selected sections. Because the reconstructed forms are presented primarily in *Hangul*, they reflect largely orthographic and literary reconstructions rather than direct phonetic transcription in IPA. Accordingly, the Korean section is not treated in the present study as primary phonetic evidence to the same extent as sections reconstructed directly in phonetic notation. Nevertheless, because *Hangul* was created during roughly the same period as HHY and because fifteenth-century Korean phonology has been studied extensively on the basis of detailed contemporary descriptions, the Korean section remains an important auxiliary resource for identifying broader transcription patterns and phonetic tendencies within HHY.

Based on the characteristics of each section and the previous scholarship reviewed above, Table 9 summarizes which features of each section are employed for comparison with the Korean section in the present study. In Table 9, the symbol '○' indicates relatively high reliability, '△' indicates relatively limited reliability, and '×' indicates that the material is considered too unreliable to be included in the analysis.

**Table 9** Reliability of previous reconstructions as references for reconstructing the HHY transcription system in onset, nucleus, and coda

| | Korean | Japan. | Viet. | Tibetan | Uyghur | Mongol. | Persian | Malay | Cham |
|---|---|---|---|---|---|---|---|---|---|
| **Onset** | △ | ○ | ○ | ○ | ○ | ○ | ○ | △ | △ |
| **Nucleus** | △ | ○ | × | ○ | ○ | × | ○ | △ | △ |
| **Coda** | △ | ○ | × | ○ | ○ | ○ | ○ | △ | △ |

Meanwhile, as the selected reconstructions were produced using different transcription conventions, most of them were not presented in IPA, with the exception of the Japanese, Tibetan, and Vietnamese sections, whose reconstructions were therefore adopted as presented in the original secondary sources. For the remaining sections, non-IPA symbols were systematically converted into their corresponding IPA symbols on the basis of established phonological studies of each language. Table 10 summarizes the correspondences adopted in this conversion:

**Table 10** Correspondences adopted in converting non-IPA symbols into IPA across the selected reconstructions

| IPA | Uyghur | Mongol. | Persian | Malay | Cham |
|---|---|---|---|---|---|
| j | y | y | y | y | y |
| w | | | | | u, v |
| tɕ | č | č | č | ch | č |
| dʑ | ǰ | ǰ | ǰ | j | ǰ |
| ʃ | š | š | š | | ç |
| r | r | r | r | | |
| ʐ | ž | | ž | | |
| z | ẓ | | | | |
| ŋ | | | | ng | ng |
| ɲ | | | | | ñ |
| ɛ | | | | | ai |
| ə | | | | ĕ | ĕ |
| ɔ | | | | | au |
| ø | | | | | ö |
| œ | | | | | o' |
| ʰ | | | | h | h |
| ː | | | - | - | - |

In some cases, the phonetic value of symbols used in the original reconstructions remains ambiguous. For example, the symbol 'r' does not always clearly distinguish between [ɾ] and [r], and the symbol 'a' may correspond to either [a] or [ɑ]. Such ambiguities were retained rather than arbitrarily resolved, and their potential impact is considered in the subsequent analysis. The revised and converted reconstruction was then parsed to be aligned with every Chinese character of HHY. Mostly either a syllable or a phoneme was aligned with a Chinese character, while more than one syllable was also rarely aligned with a character. Finally, the parsed reconstruction was manually digitized and aligned with HHY into a single spreadsheet.

## 3.3 Digitization and phonological encoding of *Huìtóngguǎnxì Huáyíyìyǔ*

For the purposes of systematic analysis, HHY was digitized and structured as a searchable dataset, with only the second row of each entry manually entered into an Excel spreadsheet. This choice was made for two main reasons. First, while the third row is absent from a substantial number of language sections, the second row is consistently attested across all sections included in HHY. Second, the second row contains Chinese transcriptions of lexical items or phrases spoken in the target language rather than metalinguistic or auxiliary information, making it the most direct and comparable source for phonological analysis.

Differences in character form or ordering across extant manuscripts were resolved by selecting a single representative form on the basis of corrections proposed in previous studies or through comparison between the reconstructed target-language form and the phonetic values of the candidate Chinese characters. Throughout the digitization process, the oldest available manuscript was treated as the primary reference in order to minimize variation introduced by later editions.

When a selected character was not supported by Unicode, the closest available character with an equivalent phonetic value was substituted and enclosed in square brackets. Each transcription character was then entered as a separate row in the dataset, while additional columns were used to assign indexical identifiers and align each character with information on Chinese phonology and its IPA reconstruction. The indexical identifiers encode both the order of lexical entries in the original text and the position of each character within an entry (A: word-initial, B: word-medial, C: word-final), thereby allowing positional effects in transcription practices to be examined systematically.

Each transcription character was subsequently aligned with Chinese phonological categories appropriate to the fifteenth and sixteenth centuries. The *yīnxì* 音系 of all characters was first encoded according to *zhōngyuányīnyùn* 中原音韻 and then adjusted to reflect phonological developments of the Late *Míng* 明 period on the basis of Tōdō (1978) and Kwon (1995). *shēngmǔ* 聲母 categories were encoded using traditional category names, whereas *yùnmǔ* 韻母 categories were represented in Romanized form. This asymmetry reflects methodological considerations: while the *yīnxì* 音系 of *shēngmǔ* 聲母 not explicitly discussed in Tōdō (1978) can generally be inferred with relative consistency from Middle Chinese phonology, developments affecting *yùnmǔ* 韻母 are less regular and require closer examination. Accordingly, *yùnmǔ* 韻母 values were reconstructed with reference to both Tōdō (1978) and Kwon (1995) in order to approximate fifteenth-century Chinese as closely as possible. For example, *kāiyīn* 開音 and *yùnwěi* 韻尾, represented as /i/ and /u/ in Tōdō (1978), were encoded as /j/ and /w/, respectively, following Kwon (1995).

# 4 Cross-linguistic Analysis of *Huìtóngguǎnxì Huáyíyìyǔ* Transcription

The reliability of previous reconstructions varies substantially across the language sections, and the correspondences observed in HHY cannot therefore be treated as equally secure in all cases. Some are supported by well-established historical evidence and internally coherent phonological systems, whereas others depend more heavily on isolated forms, modern reflexes, or uncertain secondary interpretations. The present study accordingly does not treat all attested correspondences as equally reliable when reconstructing the transcriptional principles underlying HHY.

Correspondences were evaluated selectively, not only in terms of frequency and cross-linguistic distribution, but also with respect to the historical plausibility of the reconstructed forms on which they depend. Exceptional correspondences were not excluded simply because they were infrequent or irregular. Exclusion was instead limited to cases in which the underlying reconstruction lacked independent historical support, relied excessively on modern forms, or conflicted with established diachronic evidence.

## 4.1 Distribution of Main Transcription

### 4.1.1 *Shēngmǔ* 聲母

The MT distribution in onset position shows a high degree of regularity across the HHY corpus. Across multiple language sections, *shēngmǔ* 聲母 transcription patterns are used in consistent ways that cannot be reduced to the phonological system of any single target language. Table 12 on the following page provides a preliminary working summary of *shēngmǔ* 聲母 transcription patterns across the eight language sections prior to selective evaluation. One exception concerns the alveolar-series *shēngmǔ* 聲母 in the Vietnamese section. These patterns are excluded from Table 12 because they are not directly comparable to the corresponding patterns observed in the other language sections, as shown in Table 11

**Table 11** Exceptional patterns of *shēngmǔ* 聲母 in the Vietnamese section

| **Shengmu** | *zhào* 照 | *jīng* 精 | *rì* 日 | *xīn* 心 | *shěn* 審 | *qīng* 清 | *chuān* 穿 |
|---|---|---|---|---|---|---|---|
| **Sound** | ʧj, tʂ, z | ʧj, t, ʂ, z | z, ɖ, ɲ | ʂ, $t^h$ | ʂ, $t^h$ | s | ʂ |

This exceptional distribution can be attributed to diachronic developments in Vietnamese phonology. In particular, Vietnamese /t/ derives historically from /*s/, while Sino-Vietnamese /čy/ reflects multiple *shēngmǔ* 聲母 categories, including *jīngmǔ* 精母, *zhàomǔ* 照母, and *qīngmǔ* 清母. As a result, the Vietnamese section does not provide a reliable basis for identifying general HHY transcription patterns for alveolar *shēngmǔ* 聲母 and is therefore excluded from the reconstruction of phonetic values for the alveolar group.

Table 12 is organized as follows. The symbols shown in the upper part of each cell largely follow IPA conventions, although 'ɳ̞' is not an official IPA symbol. The lower part of each cell lists the *shēngmǔ* 聲母 used for transcription without *mǔ* 母 removed for brevity. When multiple *shēngmǔ* 聲母 correspond to a single phonetic value, all relevant categories are included. The post-uvular stop /q/ is also transcribed using *jiànmǔ* 見母 in the Uyghur section and Persian sections. However, this correspondence is omitted from Table 12 for the sake of a more conciseness. As discussed in **A** below, the *jiànmǔ* 見母 transcriptions in the Uyghur and Persian sections are not ultimately retained in the reconstruction.

Among the *shēngmǔ* 聲母 categories, *míngmǔ* 明母, *nímǔ* 泥母, *láimǔ* 來母, *xiǎomǔ* 曉母, and *fēimǔ* 非母 show highly stable transcription patterns that can be accepted without selective filtering. In each case, the corresponding transcriptional domain is internally continuous, and no evidence suggests that phonetic values within these domains were systematically transcribed using other *shēngmǔ* 聲母 categories. These stable domains are indicated in bold in the table and may be summarized as follows: *míngmǔ* 明母 transcribes bilabial nasals; *nímǔ* 泥母 transcribes nasals articulated from the alveolar to the palatal region; *láimǔ* 來母 transcribes liquid consonants; *xiǎomǔ* 曉母 transcribes fricatives articulated from the velar to the glottal region; and *fēimǔ* 非母 transcribes voiceless labial fricatives.

For the remaining *shēngmǔ* 聲母 categories, selective acceptance is necessary. The primary criterion adopted here is the reliability of the evidence presented in previous studies on each language section, particularly whether the proposed reconstructions are supported by

**Table 12** Working summary of *shēngmŭ* 聲母 transcription patterns across the eight language sections

<table>
<tr><td colspan="3" rowspan="2"></td><td colspan="2">chun 脣</td><td colspan="6">she 舌, chi 齒</td><td>ya 牙</td><td>hou 喉</td></tr>
<tr><td>Bilabial</td><td>Labio-dental</td><td>Dental</td><td>Alveolar</td><td>Post-alveolar</td><td>Alveo-palatal</td><td>Palatal</td><td>Retroflex</td><td>Velar</td><td>Glottal</td></tr>
<tr><td rowspan="4">stop</td><td rowspan="2">Voiceless</td><td>Unaspirated</td><td>p<br>幇滂</td><td></td><td></td><td colspan="2">t<br>端透</td><td>ȶ<br>精</td><td></td><td></td><td>k<br>見溪</td><td></td></tr>
<tr><td>Aspirated</td><td>pʰ<br>滂</td><td></td><td></td><td colspan="2">tʰ<br>透</td><td></td><td></td><td></td><td>kʰ<br>溪</td><td></td></tr>
<tr><td rowspan="2">Voiced</td><td>Unaspirated</td><td>b<br>幇</td><td></td><td></td><td colspan="2">d<br>端</td><td>ȡ<br>精</td><td></td><td>ɖ<br>端</td><td>g<br>見</td><td></td></tr>
<tr><td>Aspirated</td><td></td><td></td><td></td><td colspan="2"></td><td></td><td></td><td></td><td>gʰ<br>見</td><td></td></tr>
<tr><td rowspan="4">Affricate</td><td rowspan="2">Voiceless</td><td>Unaspirated</td><td></td><td></td><td></td><td>ʦ<br>精</td><td>ʧ<br>照清穿</td><td>ʨ<br>精照</td><td></td><td>tʂ<br>照</td><td></td><td></td></tr>
<tr><td>Aspirated</td><td></td><td></td><td></td><td></td><td></td><td>ʨʰ<br>穿</td><td></td><td>tʂʰ<br>穿</td><td></td><td></td></tr>
<tr><td rowspan="2">Voiced</td><td>Unaspirated</td><td></td><td></td><td></td><td>ʣ<br>精</td><td>ʤ<br>照</td><td>dʑ<br>照穿</td><td></td><td>dʐ<br>照</td><td></td><td></td></tr>
<tr><td>Aspirated</td><td></td><td></td><td></td><td></td><td></td><td>dʑʰ<br>穿</td><td></td><td></td><td></td><td></td></tr>
<tr><td rowspan="2">Fricative</td><td colspan="2">Voiceless</td><td>ɸ<br>非</td><td>f<br>非</td><td></td><td>s<br>心審</td><td>ʃ<br>審</td><td>ɕ<br>心</td><td></td><td>ʂ<br>審</td><td>x<br>曉</td><td>h<br>曉</td></tr>
<tr><td colspan="2">Voiced</td><td></td><td>v<br>幇</td><td></td><td>z<br>心精</td><td>ʒ<br>日</td><td>ʑ<br>心</td><td></td><td>ʐ<br>日</td><td>ɣ<br>見影</td><td></td></tr>
<tr><td rowspan="5">Liquid</td><td colspan="2">Trill</td><td></td><td></td><td></td><td colspan="3">r<br>來</td><td></td><td></td><td></td><td></td></tr>
<tr><td colspan="2">Tap</td><td></td><td></td><td></td><td colspan="3">ɾ<br>來</td><td></td><td></td><td></td><td></td></tr>
<tr><td colspan="2">Lateral fricative</td><td></td><td></td><td colspan="4">ɬ<br>來</td><td></td><td></td><td></td><td></td></tr>
<tr><td colspan="2">Approximant</td><td></td><td></td><td colspan="4">l<br>來</td><td></td><td></td><td></td><td></td></tr>
<tr><td colspan="2">Lateral approximant</td><td></td><td></td><td colspan="4">ɺ<br>來</td><td></td><td></td><td></td><td></td></tr>
<tr><td colspan="3">Nasal</td><td>m<br>明</td><td></td><td></td><td colspan="2">n<br>泥</td><td>ȵ<br>泥</td><td>ɲ<br>泥</td><td></td><td></td><td></td></tr>
</table>

sufficient historical and phonological evidence. The specific decisions regarding acceptance and exclusion are discussed below.

### A. *jiànmǔ* 見母*–duānmǔ* 端母*–bāngmǔ* 幫母*, xīmǔ* 溪母*–tòumǔ* 透母*–pāngmǔ* 滂母

The distribution of stop-series *shēngmǔ* 聲母 across the language sections is presented in Table 13.

**Table 13** Correspondence patterns of stop-series *shēngmǔ* 聲母

| Section | *jiàn* 見 | *xī* 溪 | *duān* 端 | *tòu* 透 | *bāng* 幫 | *pāng* 滂 |
|---|---|---|---|---|---|---|
| **Korean** | k | kʰ | t | tʰ | p | pʰ |
| **Japan.** | g, k | – | t, d | – | b, p | – |
| **Vietnam.** | g, k, gʰ | kʰ | ɖ, t | tʰ | b, v | – |
| **Tibetan** | g, k | kʰ | d, t | tʰ | b, p | pʰ |
| **Uyghur** | g, q | k | d | t | b | p |
| **Mongol.** | g | k | d | t | b | b |
| **Persian** | g, q | k | d | t | b | p |
| **Malay** | g, k | – | d, t | – | b, p | – |
| **Cham** | g, k | kʰ | d, t | tʰ | b, p | pʰ |

In the Mongolian section, both *bāngmǔ* 幫母 and *pāngmǔ* 滂母 have been analyzed as being used to transcribe [b]. This interpretation in the secondary literature is based on the phonological system of Written Mongolian, which has only /b/ as a bilabial stop. Table 14 is the relevant examples presented in Ochi (2004).

**Table 14** Examples from the Mongolian section

| Index | 1st row | 2nd row | Mongolian |
|---|---|---|---|
| M–1 | 師傅 | 把黑失 | /**ba**ɣši/ (文)***ba***ɣši |
| | 三 | 忽兒班 | /ɣur***ban***/ (文)ɣur***ban*** |

However, previous reconstructions of the Middle Mongolian phonological system indicate that Middle Mongolian bilabial stops exhibited a voicing contrast (Rybatzki 2003, p. 64). In light of this evidence, the reconstruction of bilabial stops proposed in Ochi (2004) requires reconsideration. A systematic reanalysis, however, is difficult because Ochi (2004) does not provide a complete list of entries. This issue is therefore left for future research.

Across the language sections, the transcriptional distribution of *jiànmǔ* 見母*–duānmǔ* 端母*–bāngmǔ* 幫母 and *xīmǔ* 溪母*–tòumǔ* 透母*–pāngmǔ* 滂母 falls into two major patterns. In the first pattern, found primarily in the Cham, Tibetan, and Vietnamese sections, target-language stops are distinguished according to aspiration: unaspirated stops are transcribed with *jiànmŭ* 見母–*duānmŭ* 端母–*bāngmŭ* 幇母, whereas aspirated stops are transcribed with *xīmŭ* 溪母–*tòumŭ* 透母–*pāngmŭ* 滂母. The Vietnamese section, however, is excluded from further consideration because the Vietnamese reflected in HHY lacks voiceless bilabial stops, limiting its usefulness for identifying general transcriptional patterns of stop consonants.

Second, when the target language lacks an aspirated stop series, two additional subpatterns are observed. In the first, voiced stops are transcribed with *jiànmŭ* 見母–*duānmŭ* 端母–*bāngmŭ* 幇母, whereas voiceless stops are transcribed with *xīmŭ* 溪母–*tòumŭ* 透母–*pāngmŭ* 滂母. This pattern is found in the Mongolian, Uyghur, and Persian sections. In the second subpattern, both voiced and voiceless stops are transcribed with *jiànmŭ* 見母–*duānmŭ* 端母–*bāngmŭ* 幇母. This pattern is observed in the Korean, Malay, and Japanese sections.

### B. *xīnmŭ* 心母–*qīngmŭ* 清母, *shĕnmŭ* 審母–*chuānmŭ* 穿母

Table 15 presents the cross-linguistic distribution of fricative and affricate *shēngmŭ* 聲母 across the analyzed language sections.

**Table 15** Correspondence patterns of fricatives and affricates

| Section | *xīn* 心 | *shĕn* 審 | *qīng* 清 | *chuān* 穿 |
|---|---|---|---|---|
| **Korean** | s | ʃ | tsʰ | ʧʰ |
| **Japan.** | s | ʃ | – | – |
| **Tibetan** | s, z, ɕ, ʑ | ʂ | ʧ | ʨʰ, ʈʂʰ |
| **Uyghur** | s | ʃ | – | ʧ |
| **Mongol.** | s | ʃ | – | ʧ |
| **Persian** | s | ʃ | – | ʧ |
| **Malay** | s | – | – | – |
| **Cham** | s | s | – | ʥ, ʥʰ |

Several observations can be made regarding the transcription patterns presented in Table 15. First, the transcription pattern of *xīnmŭ* 心母 in the Tibetan section differs from the patterns observed in the other language sections. Second, the transcription patterns of *shĕnmŭ* 審母 and *chuānmŭ* 穿母 in the Cham section likewise diverge from the general cross-linguistic pattern. Third, *qīngmŭ* 清母 is only rarely used in transcription.

First, the basis for the claim in Nishida (1963) that *xīnmŭ* 心母 in the Tibetan section was used to transcribe not only /s/ but also /z/, /ɕ/, and /ʑ/ remains unclear. These values appear to have been inferred primarily from Written Tibetan. Moreover, /z/ has already merged with /s/ in the modern Lhasa dialect, while /ɕ/ and /ʑ/ are not directly attested as independent phonemes in modern Tibetan dialects (Beyer 1992). In the present study, priority is therefore given to the transcription patterns observed in the other language sections, and the proposal that *xīnmŭ* 心母 in the Tibetan section also represented /z/, /ɕ/, and /ʑ/ is not adopted.

Next, the only study that reconstructs the Cham section, Edwards and Blagden (1939), largely reflects modern Cham usage, while its principal source, Aymonier and Cabaton (1906), is itself a twentieth-century Cham dictionary. As a result, the phonetic value of forms transcribed with ‘s’ must be inferred indirectly through comparison with Proto-Chamic reconstructions and modern dialect evidence. On this basis, the s transcribed with *shěnmǔ* 審母 may originally have been closer to [ʃ] in pronunciation. The Cham section is therefore excluded from the interpretation of the transcriptional pattern associated with *shěnmǔ* 審母.

Finally, if the symbols reconstructed in Edwards and Blagden (1939) are interpreted literally, *chuānmǔ* 穿母 in the Cham section would appear to transcribe /j/ and /jh/. Following the description in Aymonier and Cabaton (1906) that /j/ corresponds to Serbo-Croatian /đ/ and that /jh/ represents a more strongly aspirated variant of /j/, these symbols are interpreted here as /dʑ/ and /dʑ$^h$/, respectively. However, as noted above, Edwards and Blagden (1939) largely reflects modern Cham without an explicit historical-linguistic framework, while Aymonier and Cabaton (1906) is likewise a twentieth-century dictionary. Since the reconstructed values /dʑ/ and /dʑ$^h$/ diverge substantially from the transcriptional patterns associated with *chuānmǔ* 穿母 in the other language sections, they are not adopted in the present analysis.

### C. *jīngmǔ* 精母–*zhàomǔ* 照母–*rìmǔ* 日母

**Table 16** Correspondence patterns of the rest

| Section | *jīng* 精 | *zhào* 照 | *rì* 日 |
|---|---|---|---|
| **Korean** | ʦ, z | ʧ | ʒ |
| **Japan.** | ʦ, ʣ | ʧ | ʒ |
| **Tibetan** | ʈ, ɖ, ʦ, ʣ | tʂ, dʐ, ʨ, ʥ | ʐ |
| **Uyghur** | z | ʤ | ʒ |
| **Mongol.** | z | ʤ | – |
| **Persian** | z | ʤ | – |
| **Malay** | – | ʧ, ʤ | – |
| **Cham** | ʨ | dʑ | – |

Two observations can be made regarding the transcription patterns presented in Table 16. First, the transcription patterns associated with *jīngmǔ* 精母 and *zhàomǔ* 照母 in the Tibetan section differ from those observed in the other language sections. Second, the transcription pattern associated with *jīngmǔ* 精母 in the Cham section likewise diverges from the general cross-linguistic pattern.

First, the analysis proposed in Nishida (1963), according to which *jīngmǔ* 精母 in the Tibetan section was used to transcribe /ʈ/ and /ɖ/, while *zhàomǔ* 照母 was used to transcribe /ʨ/ and /ʥ/, is not adopted in the present study. These values appear to have been inferred primarily from Written Tibetan. In modern Tibetan dialects, however, such reflexes are largely confined to the Lhasa dialect and are otherwise reflected differently across dialects (Beyer 1992). Although such correspondences might be considered if comparable transcription patterns were attested in other language sections, the evidence presented in Nishida (1963) alone is insufficient to establish correspondences between *jīngmǔ* 精母 and /ʈ, ɖ/ or between *zhàomǔ* 照母 and /ʨ, ʥ/.

The Cham section might appear to support a correspondence between *zhàomǔ* 照母 and /ʥ/, but the reconstructions proposed in Edwards and Blagden (1939) are difficult to accept. The source on which Edwards and Blagden (1939) relies, Aymonier and Cabaton (1906), explains the phonetic values of its symbols primarily through comparison with Serbo-Croatian rather than through explicit phonetic description. For example, the symbol ‘c’ in Edwards and Blagden (1939), reconstructed here as /ʨ/, is described only as a strongly palatalized prepalatal sound, intermediate between the ‘ti’ in French *tiare* and the ‘qui’ in *inquiet*; cf. Serbo-Croatian /ć/. On this basis, it is difficult to determine whether the intended value was closer to [ʨ] or [ʧ]. As in the Tibetan case, such correspondences might be considered if they were independently supported by evidence from other language sections. However, the currently available evidence does not provide sufficient grounds for accepting that *jīngmǔ* 精母 transcribed /ʨ/ or that *zhàomǔ* 照母 transcribed /ʥ/.

The transcription patterns retained after selective evaluation are summarized in Table 17. Patterns rejected in the preceding discussion have been removed from Table 12, while those judged reliable have been retained. Table 17 therefore presents the final cross-linguistic summary of *shēngmǔ* 聲母 transcription patterns across the eight language sections.

### 4.1.2 *yùnmǔ* 韻母

The MT distribution in rime position likewise shows a high degree of regularity across the HHY corpus. Across multiple language sections, *yùnmǔ* 韻母 transcription patterns are employed in consistent ways that cannot be reduced to the phonological system of any single target language. Table 18 provides a preliminary working summary of *yùnmǔ* 韻母 transcription patterns across the eight language sections prior to selective evaluation.

**Table 18** Working summary of *yùnmǔ* 韻母 transcription patterns across the eight language sections

| Section | /a/ | /ə/ | /jə/ | /ɨ/ | /jɨ/ | /wo/ | /u/ | /-n/ | /-ŋ/ |
|---|---|---|---|---|---|---|---|---|---|
| **Korean** | a, ʌ | ʌ,ɨ, ə | i | ɨ | i | o | u | n | ŋ |
| **Japanese** | a | a, o | i | i, u | i | o | u | n | ŋ |
| **Tibetan** | a | a, o | i | i | i | o | u | n | ŋ |
| **Uyghur** | a | ä, a | i | – | i | o | u | n | ŋ |
| **Persian** | a, aː | a | i | – | i | o | u, uː | n | ŋ |
| **Malay** | a | ə, o | i | i | i | – | u | n | ŋ |
| **Cham** | a | ɔw, a | i | – | i, iː | – | u | n | ŋ |

The *yùnmǔ* 韻母 /jə/, /jɨ/, /wo/, and /u/, as well as the codas /-n/ and /-ŋ/, show stable transcription patterns that do not require selective evaluation. The *yùnmǔ* 韻母 /jə/ and /jɨ/ are consistently used to transcribe the front high vowel [i]. The *yùnmǔ* 韻母 /wo/ is used to transcribe the back mid vowel [o], while /u/ is used for the back high vowel [u]. Finally, the *yùnwěi* 韻尾 /-n/ and /-ŋ/ correspond consistently to [n] and [ŋ], respectively.

**Table 17** Working summary of *shēngmǔ* 聲母 transcription patterns across the eight language sections

| | | | chun 脣 | | she 舌, chi 齒 | | | | | | ya 牙 | hou 喉 |
|---|---|---|---|---|---|---|---|---|---|---|---|---|
| | | | Bilabial | Labio-dental | Dental | Alveolar | Post-alveolar | Alveo-palatal | Palatal | Retroflex | Velar | Glottal |
| stop | Voiceless | Unaspirated | p 幇 | | | t 端 | | | | | k 見 | |
| | | Aspirated | pʰ 滂 | | | tʰ 透 | | | | | kʰ 溪 | |
| | Voiced | Unaspirated | b 幇 | | | d 端 | | | | | g 見 | |
| | | Aspirated | | | | | | | | | | |
| Affricate | Voiceless | Unaspirated | | | | ts 精 | tʃ 照 | | | tʂ 照 | | |
| | | Aspirated | | | | | | | | tʂʰ 穿 | | |
| | Voiced | Unaspirated | | | | ʣ 精 | ʤ 照 | | | dʐ 照 | | |
| | | Aspirated | | | | | | | | | | |
| Fricative | Voiceless | | ɸ 非 | f 非 | | s 心 | ʃ 審 | | | ʂ 審 | x 曉 | h 曉 |
| | Voiced | | | v 幇 | | z 精 | ʒ 日 | | | ʐ 日 | ɣ 見影 | |
| Liquid | Trill | | | | | r 來 | | | | | | |
| | Tap | | | | | ɾ 來 | | | | | | |
| | Lateral fricative | | | | ɬ 來 | | | | | | | |
| | Approximant | | | | l 來 | | | | | | | |
| | Lateral approximant | | | | ɺ 來 | | | | | | | |
| Nasal | | | m 明 | | | n 泥 | | ȵ 泥 | ɲ 泥 | | | |

By contrast, the transcription patterns associated with the *yùnmǔ* 韻母 /a/, /ə/, and /ï/ require further examination. Across the language sections, vowels transcribed with /a/ are consistently represented as ‘a’, but their precise phonetic values cannot be determined with certainty. Nevertheless, the secondary sources consulted in the present study make it clear that these vowels belonged broadly to the low-vowel range.

**A. *yùnmǔ* 韻母 /a/ - /ə/**

**Table 19** Distribution of vowel correspondences across language sections

| Section | ə | a | o | u | ä | ɔw | Total |
|---|---|---|---|---|---|---|---|
| **Japan** | - | 63 | 55 | 13 | - | - | 144 |
| **Tibetan** | | 85 | 93 | 26 | | | 238 |
| **Uyghur** | | 71 | 26 | 1 | 81 | | 242 |
| **Persian** | | 230 | - | 8 | - | | 277 |
| **Malay** | 45 | 3 | 18 | 7 | | | 87 |
| **Cham** | - | 37 | - | 9 | | 45 | 126 |

In Table 19, Total refers to the number of transcription characters excluding ST and unresolved items. The table does not present all attested correspondence patterns, but only the two most frequent correspondences for each category. An exception is made for /u/, which is included because it appears consistently across the transcription patterns of all language sections.

The most noteworthy pattern in Table 19 is that observed in the Malay section. The *yùnmǔ* 韻母 /ə/ was used not only to transcribe the vowel /ə/ in Malay, but also in a near one-to-one correspondence with it. Although /ə/ was occasionally used to transcribe /o/ as well, this correspondence is of lesser significance, since /o/ does not appear to have belonged to the phonemic inventory of native Malay vocabulary. The Malay data therefore strongly suggest that the transcriptional value of the *yùnmǔ* 韻母 /ə/ was close to [ə].

This perspective also helps narrow the transcriptional range associated with the *yùnmǔ* 韻母 /a/. In the Japanese, Tibetan, and Uyghur sections, vowels transcribed with the *yùnmǔ* 韻母 /a/ were also occasionally represented using the *yùnmǔ* 韻母 /ə/. Moreover, the Chinese rime /ə/ was used not only for [a], but also for relatively back vowels such as [o], [u], and [ä]. Since the transcriptional value of the *yùnmǔ* 韻母 /ə/ has been identified as [ə], the transcriptional range of /a/ must have been located further back in vowel space.

Among the vowel correspondences, the transcription patterns observed in the Japanese section are particularly informative. Although the *yùnmǔ* 韻母 /ï/ does not show a consistent one-to-one correspondence with either [i] or [u], it was used to transcribe both vowels. This suggests that its transcriptional value occupied an intermediate position between the two.

It should also be noted that many language sections could not be incorporated into the present discussion. This does not mean, however, that the *yùnmǔ* 韻母 /ï/ was entirely absent from their transcription systems. Rather, these sections are marked only with –’ because the *yùnmǔ* 韻母 /ï/ appears exclusively in ST rather than in the main transcription layer. A similar tendency can be observed in the Tibetan section, where *yùnmǔ* 韻母 /ï/ occurs more frequently in ST than in the transcription of i' itself. These ST-only usages likely reflect meaningful aspects of the HHY transcription system, although a full discussion of their implications falls beyond the scope of the present study.

**Table 20** Distribution of front vowel correspondences across language sections

| Section | i | u | Total |
|---|---|---|---|
| **Japan** | 44 | 28 | 73 |
| **Tibetan** | 37 | - | 37 |
| **Uyghur** | - | | 5 |
| **Persian** | - | | 3 |
| **Malay** | - | | 16 |
| **Cham** | - | | 3 |

Based on the findings presented above, Figure 1 summarizes the phonetic values or approximate phonetic ranges associated with each *yùnmǔ* 韻母 in the HHY transcription system.

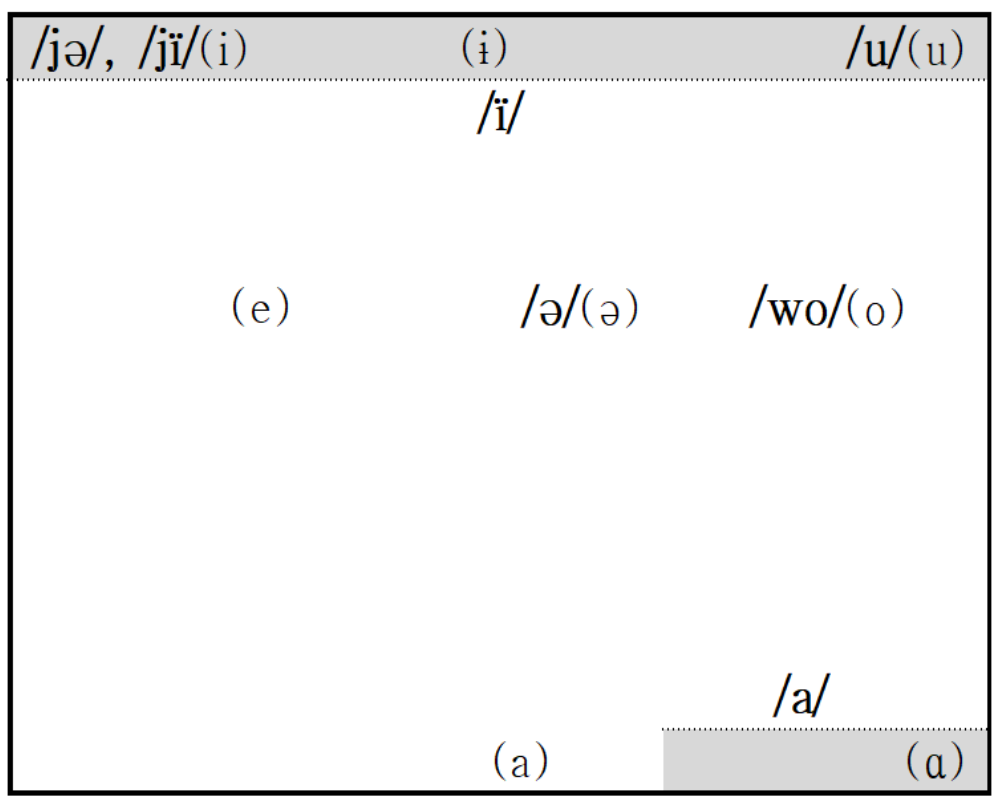

**Fig. 1** Approximate phonetic ranges of the Chinese *yunmu* in HHY transcription

The shaded area in the upper part of Figure 1 represents the transcriptional range associated with the *yùnmǔ* 韻母 /ï/, while the shaded area in the lower part represents the range associated with /a/. The symbols in parentheses - (i, ɨ, u, e, ə, o, a, ɑ) - are IPA symbols indicating the phonetic values corresponding to each position in the figure.

## 4.2 Distribution of Supplementary Transcription

A comprehensive summary of the ST patterns observed across the eight language sections is presented in Table 21. As noted earlier, the Vietnamese section is considered only with respect to MT patterns in *shēngmǔ* 聲母 transcription and is therefore excluded from the present discussion.

**Table 21** Summary of *shēngmŭ* 聲母 transcription patterns

<table>
<tr><th colspan="2" rowspan="2"></th><th colspan="2">chun 脣</th><th colspan="4">she 舌, chi 齒</th><th colspan="2">ya 牙</th><th>hou 喉</th></tr>
<tr><th>Bilabial</th><th>Labio-dental</th><th>Dental</th><th>Alveolar</th><th>Post-alveolar</th><th>Alveo-palatal</th><th>Velar</th><th>Postvelar</th><th>Glottal</th></tr>
<tr><td rowspan="2">**stop**</td><td>**Voiceless**</td><td>p<br>幇滂</td><td></td><td></td><td>t<br>端透</td><td></td><td></td><td>k<br>溪見</td><td>q<br>溪見</td><td></td></tr>
<tr><td>**Voiced**</td><td>b<br>幇</td><td></td><td></td><td>d<br>端</td><td></td><td></td><td>g<br>見</td><td></td><td></td></tr>
<tr><td rowspan="2">**Affricate**</td><td>**Voiceless**</td><td></td><td></td><td></td><td></td><td>ʧ<br>穿照</td><td></td><td></td><td></td><td></td></tr>
<tr><td>**Voiced**</td><td></td><td></td><td></td><td></td><td>ʤ<br>照日</td><td></td><td></td><td></td><td></td></tr>
<tr><td rowspan="2">**Fricative**</td><td>**Voiceless**</td><td></td><td>f<br>非</td><td></td><td>s<br>心</td><td>ʃ<br>審</td><td></td><td>x<br>曉</td><td></td><td>h<br>曉</td></tr>
<tr><td>**Voiced**</td><td></td><td></td><td></td><td>z<br>精</td><td>ʒ<br>日</td><td></td><td>ɣ<br>影溪見曉</td><td></td><td></td></tr>
<tr><td rowspan="3">**Liquid**</td><td>**Trill**</td><td></td><td></td><td></td><td colspan="3">r<br>兒來</td><td></td><td></td><td></td></tr>
<tr><td>**Tap**</td><td></td><td></td><td></td><td colspan="3">ɾ<br>兒來</td><td></td><td></td><td></td></tr>
<tr><td>**Lateral approximant**</td><td></td><td></td><td colspan="4">l<br>來兒</td><td></td><td></td><td></td></tr>
<tr><td>**Nasal**</td><td></td><td>m<br>明</td><td></td><td></td><td>n<br>/-n/</td><td></td><td></td><td></td><td></td><td></td></tr>
</table>

The basic structure of Table 21 follows that of the preceding table summarizing *shēngmǔ* 聲母 transcription patterns. Several differences should nevertheless be noted. First, *ér* 兒 refers not to the transcription character 兒 itself, but to the broader category of *érhuà* 兒化 rimes. Second, /-n/ refers to ST characters containing the *yùnwěi* 韻尾 /-n/. Although the table is based on a complete analysis of all relevant items, isolated forms that appear to reflect scribal or transcriptional errors were excluded in order to present the overall patterns more clearly.

HHY also appears to have employed a partially distinct set of transcription characters specifically for ST. Table 22 summarizes the transcription characters used for ST in each language section.

**Table 22** ST characters by *shēngmǔ* 聲母 category

| Category | ST characters |
|---|---|
| ***bāngmǔ* 幫母** | 卜, 補, 不, 白, 絣 |
| ***pāngmǔ* 滂母** | 批 |
| ***duānmǔ* 端母** | 的, 答, 得, 都 |
| ***tòumǔ* 透母** | 忒, 剔, 惕, 禿, 帖 |
| ***jiànmǔ* 見母** | 格, 革, 吉, 艮, 故, 果 |
| ***xīmǔ* 溪母** | 克, 闊, 乞, 苦 |
| ***fēimǔ* 非母** | 夫, 伏 |
| ***xīnmǔ* 心母** | 思, 習, 糸, 西, 速, 桑 |
| ***shěnmǔ* 審母** | 失 |
| ***jīngmǔ* 精母** | 子, 聚, 則 |
| ***rìmǔ* 日母** | 日 |
| ***chuānmǔ* 穿母** | 赤, 除, 出 |
| ***zhàomǔ* 照母** | 只, 褚 |
| ***xiǎomǔ* 曉母** | 黑, 誠, 哈, 蛤, 吸 |
| ***yǐngmǔ* 影母** | 額, 兒 |
| **erhuayun 兒化韻** | 兒, 二 |
| ***láimǔ* 來母** | 力, 勒, 里, 剌, 魯, 綠, 羅, 路, 弄, 利 |
| ***míngmǔ* 明母** | 密, 母, 木 |
| **/-n/** | 音 |

However, not all transcription characters listed in Table 22 can be regarded as genuine ST characters. As discussed above, some forms occur only sporadically and do not appear to reflect systematic ST usage within the HHY transcription system. It is therefore necessary to distinguish such cases from transcription characters that were consistently used for ST across the corpus. In the present study, ST characters were identified on the basis of two criteria.

First, as the strongest criterion, we examined whether a given transcription character was used for ST in more than one language section. Since each language section was compiled by a different bureau (*guǎn* 館), the sections were likely prepared by different compilers or working groups. Although no direct records of the HHY compilation process survive, this can reasonably be inferred from the officials associated with each bureau as listed in the London manuscript. If the same transcription character was independently used for ST across language sections compiled by different groups, it is reasonable to assume that the character functioned as an established ST character at the time.

Second, when a transcription character appeared in only a single language section, we examined whether it occurred in more than one entry. Even when a character appears in multiple entries, if all instances involve the transcription of the same morpheme, the pattern is better explained as a consequence of HHY's tendency toward orthographic consistency rather than as evidence that the character functioned as a dedicated ST character.

**A. *jiànmŭ* 見母 - *xīmŭ* 溪母 - *yĭngmŭ* 影母 - *xiăomŭ* 曉母**

*Jiànmŭ* 見母 and *xīmŭ* 溪母 were primarily used to represent ST velar stops. However, together with *yĭngmŭ* 影母 and *xiăomŭ* 曉母, they were also used in the transcription of ST [ɣ]. For this reason, all four categories are considered together here. Table 23 summarizes the distributional patterns of the relevant transcription characters, irrespective of the specific phonemes or phones they represented.

**Table 23** Distribution of selected ST characters across language sections

| | | Persian | Uyghur | Mongolian | Cham | Total |
|---|---|---|---|---|---|---|
| ***jiàn* 見** | 格 | | | | 1 | 1 |
| | 革 | 27 | 1 | | | 28 |
| | 吉 | 1 | | | | 1 |
| | 艮 | | 1 | | | 1 |
| | 故 | | 1 | | | 1 |
| | 果 | | 1 | | | 1 |
| ***xī* 溪** | 克 | 22 | 42 | ⊙ | | 64+⊙ |
| | 闊 | | 2 | | | 2 |
| | 乞 | | 1 | | | 1 |
| | 苦 | 1 | 3 | | | 4 |
| ***yĭng* 影** | 額 | 11 | | | | 11 |
| | 兒 | | 2 | | | 2 |
| ***xiăo* 曉** | 黑 | | 18 | ⊙ | | 18+⊙ |
| | 蛤 | | | ⊙ | | ⊙ |
| | 虩 | 49 | | | | 49 |
| | 哈 | 1 | | | | 1 |
| | 吸 | 1 | | | | 1 |

According to the criteria proposed above, the only *jiànmŭ* 見母 character that can be identified as an ST character is *gé* 革. Within the *xīmŭ* 溪母 group, *kè* 克 likewise qualifies as an ST character. *Kŭ* 苦, however, should not be excluded entirely, since it occurs in more than one language section, albeit only in a limited number of examples.

As for *ér* 兒, it was used primarily in the Uyghur section to represent [r] in forms that had already undergone *érhuà* 兒化. Accordingly, the only *yǐngmǔ* 影母 character that can be regarded as an ST character is *é* 額. Within the *xiǎomǔ* 曉母 group, *hēi* 黑 and *xià* 諕 qualify as ST characters. By contrast, *gé* 蛤 was excluded because the full entries of the Mongolian section cannot be verified, making it difficult to determine whether the character was systematically used across multiple entries. Moreover, it is not attested in any other language section.

Table 24 summarizes the distributional patterns of *jiànmǔ* 見母, *xīmǔ* 溪母, *yǐngmǔ* 影母, and *xiǎomǔ* 曉母 when used as ST characters across the language sections.

**Table 24** Correspondence patterns of *jiàn* 見, *xī* 溪, *yǐng* 影, and *xiǎo* 曉

| | Persian | Uyghur | Mongolian |
|---|---|---|---|
| ***jiàn* 見母** | g, q, | | – |
| ***xī* 溪母** | k | k, q, | k |
| ***yǐng* 影母** | | – | – |
| ***xiǎo* 曉母** | h, x | x | |

None of the ST targets shown in Table 24 involve unreleased stops. The only language sections in which the voiceless sounds [k] and [q] were represented through ST are the Persian and Uyghur sections, and in both languages these sounds did not undergo unreleased realization even in word-final position.

The fact that [q] and [ɣ] were represented differently across the language sections suggests that they were phonetically distinct. However, little is known about their precise phonetic values in word-final position in fifteenth-century Persian, Uyghur, and Mongolian. In the case of [ɣ], previous studies have shown that its phonemic status in fifteenth-century Uyghur and Mongolian was not firmly established even in onset position, and that it occurred only under specific phonological conditions (Ochi 2004). The ST patterns involving [ɣ] in the Uyghur and Mongolian sections are therefore excluded from the present discussion.

In Persian, meanwhile, [q] is attested in twentieth-century Persian as the result of a merger involving the native phoneme /ɣ/ and /k/ in Arabic loanwords, although this merger had not yet become phonologized in the thirteenth century (Pisowicz 1985). This raises the possibility that forms interpreted as ST representations of [q] in the Persian section may in fact reflect /ɣ/ or /k/. The corresponding ST patterns in the Persian section are therefore likewise excluded.

**B. *bāngmǔ* 幇母–*pāngmǔ* 滂母–*fēimǔ* 非母**

*Bāngmǔ* 幇母 and *pāngmǔ* 滂母 were primarily used to represent ST bilabial stops, while *fēimǔ* 非母 was mainly associated with ST bilabial fricatives. Table 25 summarizes the distributional patterns of the relevant transcription characters, irrespective of the specific sounds they represented.

**Table 25** Distribution of ST characters associated with *bāng* 幫, *pāng* 滂, and *fēi* 非

| | | Persian | Uyghur | Tibetan | Mongolian | Cham | Korean | Total |
|---|---|---|---|---|---|---|---|---|
| ***bāng* 幫** | 卜 | 32 | 11 | 20 | | | 3 | 66 |
| | 不 | | | | ⊙ | 6 | | 6+⊙ |
| | 補 | | 9 | 1 | | | | 10 |
| | 白 | 1 | | 3 | | | | 4 |
| | 紨 | | | 1 | | | | 1 |
| | 別 | | | | | | 1 | 1 |
| ***pāng* 滂** | 批 | | 1 | | | | | 1 |
| ***fēi* 非** | 夫 | | 3 | | | | | 3 |
| | 伏 | 25 | | | | | | 25 |

According to the criteria proposed above, *bo* 卜 and *bù* 不 in the *bāngmŭ* 幫母 group qualify as ST characters. However, *bŭ* 補 and *bái* 白 should not be excluded entirely, since they occur in more than one language section, albeit only in a limited number of examples. The character 紨 may represent an erroneous form of 補, although this can only be confirmed through direct examination of the base manuscript. In the Korean section, 別 likewise appears in only a single item, and even Kwon (1998) does not treat the reconstruction with confidence. It is therefore difficult to regard it as reliable evidence for a dedicated ST character.

In the case of *pāngmŭ* 滂母, only a single example involving *pī* 批 is attested in ST position. It therefore remains unclear whether *pāngmŭ* 滂母 possessed a dedicated ST character. As for *fēimŭ* 非母, both *fū* 夫 and *fú* 伏 qualify as ST characters.

Table 26 summarizes the distributional patterns of *bāngmŭ* 幫母, *pāngmŭ* 滂母, and *fēimŭ* 非母 when used as ST characters across the language sections.

**Table 26** Correspondence patterns of *bāng* 幫 and *fēi* 非

| | Persian | Uyghur | Tibetan | Mongolian | Cham | Korean |
|---|---|---|---|---|---|---|
| ***bāng* 幫母** | b, p | b, p | b, p | b | b, p | β |
| ***fēi* 非母** | f | f | – | – | – | – |

One particularly noteworthy pattern in Table 26 concerns [p]. As discussed for [k], [g], and [q], and as will also be shown below for [d] and [t], ST representations of stop consonants generally exhibit a tendency for voiceless sounds to be represented by aspirated *shēngmŭ* 聲母 series and voiced sounds by unaspirated series. However, despite being voiceless, [p] was represented by the unaspirated *bāngmŭ* 幫母 series.

One possible explanation is that, although reconstructed phonologically as /p/, the relevant forms may in fact have been phonetically realized as [b] under particular phonological conditions. However, as shown in Table 27, there are cases in which [p] occurring in environments that do not permit voicing was nevertheless represented through *bāngmǔ* 幇母.

**Table 27** Representative examples of transcription entries

| **Index** | **1st row** | **2nd row** | **Target** |
|---|---|---|---|
| U-34 | 雲起 | 課克科卜 | kök qop |
| U-703 | 薄 | 與卜哈納 | yupqana |
| T-6 | 雲 | 卜吝 | /prin/ *lit.* sprin |
| P-1622 | 左 | 徹卜 | čap |

### C. *duānmǔ* 端母–*tòumǔ* 透母

*Duānmǔ* 端母 and *tòumǔ* 透母 were primarily used to represent ST alveolar stops. Table 28 summarizes the distributional patterns of the relevant transcription characters, irrespective of the specific sounds they represented.

**Table 28** Distribution of ST characters associated with *duān* 端 and *tòu* 透

| | | Persian | Uyghur | Malay | Cham | Japanese | Total |
|---|---|---|---|---|---|---|---|
| ***duān* 端** | 的 | | 2 | 3 | 2 | 10 | 17 |
| | 得 | 58 | | | | | 58 |
| | 都 | 1 | | | | | 1 |
| | 答 | | | | 1 | | 1 |
| ***tòu* 透** | 忒 | 26 | 27 | | | | 53 |
| | 剔 | 14 | | | | | 14 |
| | 惕 | | 6 | | | | 6 |
| | 禿 | | 1 | | | | 1 |
| | 帖 | | 1 | | | | 1 |

According to the criteria proposed above, *de* 的 and *de* 得 in the *duānmǔ* 端母 group qualify as ST characters. Within the *tòumǔ* 透母 group, *tè* 忒 is the most robustly attested ST character, while *tī* 剔 and *tì* 惕 likewise function as valid ST characters.

Table 29 summarizes the distributional patterns of *duānmǔ* 端母 and *tòumǔ* 透母 when used as ST characters across the language sections.

Table 29 Correspondence patterns of *duān* 端 and *tòu* 透

| | Persian | Uyghur | Malay | Cham | Japanese |
|---|---|---|---|---|---|
| ***duān* 端母** | d | d | t, ʧ | t | t |
| ***tòu* 透母** | t | t | – | – | – |

The cases of [t] and [ʧ] represented by *duānmŭ* 端母 in the Malay section should be excluded from discussion. The segments reconstructed as [ʧ] and [t] in Edwards and Blagden (1931), which presents the forms in Modern Malay, may originally have corresponded to different phonemes, such as /d/, in fifteenth-century Malay. According to Maris (1980, pp. 55–56), syllable-final /t/ and /d/ in Modern Malay are unreleased in coda position, but it remains unclear when this neutralization or unreleased realization emerged historically. It is therefore difficult to rule out the possibility that forms reconstructed as [t] in Edwards and Blagden (1931) may actually have corresponded to /d/ during the period represented in the Malay section. The case of [t] represented by *duānmŭ* 端母 in the Japanese section should likewise be excluded from consideration. Previous studies disagree on whether *de* 的 in these items represented ST [t]. Matsumoto and Ding (1997) interpreted these forms as representing [t], whereas Watanabe (1961, p. 8) analyzed them as corresponding to チ.

The Cham section is the only case in which the unaspirated *shēngmŭ* 聲母 category *duānmŭ* 端母 was used to represent the voiceless sound [t]. Nevertheless, this pattern cannot simply be disregarded. The relevant [t] occurs as the first element of an initial consonant cluster, and Cham initial clusters are generally understood to have arisen through vowel loss (Thurgood 1999). It is therefore likely that this segment was indeed realized as [t] in fifteenth-century Cham.

Even so, two considerations suggest that *tòumŭ* 透母 functioned as the primary ST *shēngmŭ* 聲母 for [t], whereas *duānmŭ* 端母 functioned as the primary ST *shēngmŭ* 聲母 for [d]. First, in the Cham section, no cases are attested in which [d], contrasting with [t] in voicing, is represented through ST. Second, in both the Persian and Uyghur sections, where [t] and [d] are both represented through ST, they correspond systematically to *tòumŭ* 透母 and *duānmŭ* 端母, respectively. Therefore, although the correspondence between [t] and *duānmŭ* 端母 observed in the Cham section should not be ignored, it is better treated as a secondary correspondence.

#### D. *xīnmŭ* 心母–*shěnmŭ* 審母

Table 30 summarizes the distributional patterns of the transcription characters associated with *xīnmŭ* 心母 and *shěnmŭ* 審母, irrespective of the specific sounds they represented.

Table 30 Distribution of ST characters associated with *xīn* 心 and *shěn* 審

| | | Persian | Uyghur | Tibetan | Mongolian | Malay | Korean | Total |
|---|---|---|---|---|---|---|---|---|
| ***xīn* 心** | 思 | 51 | 36 | 13 | ⊙ | 14 | 10 | 124+⊙ |
| | 習 | | 7 | | | 1 | | 8 |
| | 糸 | | 3 | | | | | 3 |
| | 西 | | 3 | | | | | 3 |
| | 速 | | 2 | | | | | 2 |
| | 桑 | 1 | | | | | | 1 |
| ***shěn* 審** | 失 | 41 | 70 | | ⊙ | | | 111+⊙ |

According to the criteria proposed above, *sī* 思 is the most robustly attested ST character in the *xīnmŭ* 心母 group. However, *xí* 習, *sī* 糸, and *sù* 速 are also retained, since they occur in different lexical items. By contrast, *xī* 西 is excluded because all of its occurrences are confined to the representation of [s] in the same lexical item, *eski* (Uyghur-150, 491, 713). For *shěnmŭ* 審母, the corresponding ST character is *shī* 失.

Table 31 summarizes the distributional patterns of *xīnmŭ* 心母 and *shěnmŭ* 審母 when used as ST characters across the language sections.

Table 31 Correspondence patterns of *xīn* 心 and *shěn* 審

| | Persian | Uyghur | Tibetan | Mongolian | Malay | Korean |
|---|---|---|---|---|---|---|
| ***xīn* 心母** | s | s | s | s | s | s |
| ***shěn* 審母** | ʃ | ʃ | – | ʃ | – | – |

The ST patterns associated with *xīnmŭ* 心母 and *shěnmŭ* 審母 are relatively straightforward. As shown in Table 31, *xīnmŭ* 心母 was used to represent [s], whereas *shěnmŭ* 審母 was used to represent [ʃ].

**E. *chuānmŭ* 穿母–*zhàomŭ* 照母, *jīngmŭ* 精母–*rìmŭ* 日母**

Table 32 summarizes the distributional patterns of the transcription characters associated with *chuānmŭ* 穿母, *zhàomŭ* 照母, *jīngmŭ* 精母, and *rìmŭ* 日母, irrespective of the specific sounds they represented.

**Table 32** Distribution of ST characters associated with *chuān* 穿, *zhào* 照, *rì* 日, and *jīng* 精

| | | Persian | Uyghur | Korean | Total |
|---|---|---|---|---|---|
| ***chuān* 穿** | 赤 | | 9 | | 9 |
| | 除 | | 5 | | 5 |
| | 出 | | 2 | | 2 |
| ***zhào* 照** | 只 | 7 | | | 7 |
| | 褚 | | 1 | | 1 |
| ***rì* 日** | 日 | 2 | | | 2 |
| ***jīng* 精** | 子 | 19 | 58 | | 77 |
| | 聚 | | 1 | | 1 |
| | 則 | 1 | | | 1 |
| | 自 | | | 2 | 2 |

According to the criteria proposed above, *chì* 赤 appears to have been the primary ST character used for *chuānmŭ* 穿母. Although *chú* 除 is excluded because all of its occurrences are confined to the same lexical item, *üč* (Uyghur-193, 238, 250, 265, 804), *chū* 出 is retained because it occurs in different lexical items. The ST characters used for *zhàomŭ* 照母, *rìmŭ* 日母, and *jīngmŭ* 精母 were *zhĭ* 只, *rì* 日, and *zĭ* 子, respectively.

Table 33 summarizes the distributional patterns of these *shēngmŭ* 聲母 across the language sections when used as ST characters.

**Table 33** Correspondence patterns of *chuān* 穿, *zhào* 照, *jīng* 精, and *rì* 日

| | Persian | Uyghur | Korean |
|---|---|---|---|
| ***chuān* 穿母** | – | ʧ | – |
| ***zhào* 照母** | ʤ | – | – |
| ***jīng* 精母** | z | z | z |
| ***rì* 日母** | ʤ, ʒ | – | – |

The ST patterns associated with *chuānmŭ* 穿母, *zhàomŭ* 照母, and *jīngmŭ* 精母 are relatively straightforward. As shown in Table 33, *chuānmŭ* 穿母 was used to represent [ʧ], *zhàomŭ* 照母 to represent [ʤ], and *jīngmŭ* 精母 to represent [z].

The case of *rìmŭ* 日母 is somewhat different, since only two examples are attested and they correspond to different sounds. Within the Persian section, however, [ʤ] is consistently represented by *zhàomŭ* 照母 across multiple entries. This suggests that [ʒ], rather than [ʤ], was the sound represented by *rìmŭ* 日母.

**F. *láimŭ* 來母–*érhuà*-rime (兒化韻)**

Table 34 summarizes the distributional patterns of the transcription characters associated with *láimŭ* 來母 and *érhuà*-rime 兒化韻, irrespective of the specific sounds they represented.

**Table 34** Distribution of ST characters associated with *lái* 來 and *ér* 兒

| | | Persian | Uyghur | Tibetan | Mongolian | Malay | Cham | Korean | Total |
|---|---|---|---|---|---|---|---|---|---|
| ***lái* 來** | 勒 | 2 | 16 | | ⊙ | | | 2 | 20 |
| | 力 | 43 | 14 | | | 1 | | | 58 |
| | 里 | | 29 | | ⊙ | | | | 29 |
| | 剌 | 1 | | | | 3 | | | 4 |
| | 魯 | | 2 | | | | | | 2 |
| | 綠 | | 1 | | | | | | 1 |
| | 羅 | | | | | 1 | | | 1 |
| | 路 | | | | | 1 | | | 1 |
| | 弄 | | | | | 1 | | | 1 |
| | 利 | | | | | 1 | | | 1 |
| ***ér* 兒** | 兒 | 146 | 249 | 124 | ⊙ | 19 | 10 | | 548+⊙ |
| | 二 | | | | ⊙ | | | 75 | 75+⊙ |

According to the criteria proposed above, the most robustly attested ST characters for *láimŭ* 來母 are *lè* 勒, *lì* 力, and *lĭ* 里. However, *là* 剌 should not be disregarded, since it was used for ST in more than one language section, albeit in a limited number of examples. The ST characters associated with *érhuà*-rime 兒化韻 are *ér* 兒 and *èr* 二. Since *èr* 二 was primarily used in the Korean section to represent ST /l/, it may reasonably be regarded as an ST character for *érhuà*-rime 兒化韻.

Table 35 below summarizes the distributional patterns of *láimŭ* 來母 and *érhuà*-rime 兒化韻 across the language sections.

**Table 35** Correspondence patterns of *lái* 來 and *érhuà* 兒化韻

| | Persian | Uyghur | Tibetan | Mongolian | Malay | Cham | Korean |
|---|---|---|---|---|---|---|---|
| ***lái* 來母** | l, ɾ | l, ɾ | l | l, ɾ | r | – | l |
| ***érhuà* 兒化韻** | ɾ, l | ɾ, l | r | ɾ, l | r | r | l |

The patterns shown in Table 35 may initially appear inconsistent, with *láimŭ* 來母 and *érhuà*-rime 兒化韻 both used to represent [l], [r], and [ɾ]. The frequency distributions, however, reveal a much clearer tendency. Of the relevant instances of [l], 78% are represented by *láimŭ* 來母, whereas 97% of the instances of [r] are represented by *érhuà*-rime 兒化韻. At the same time, representations of [l] through *érhuà*-rime 兒化韻 and of [r] through *láimŭ* 來母 are sufficiently frequent that they are better treated as secondary correspondences rather than exceptional forms.

### G. *míngmŭ* 明母, /-n/

Table 36 summarizes the distributional patterns of the transcription characters associated with *míngmŭ* 明母 and /-n/, irrespective of the specific sounds they represented.

**Table 36** Distribution of ST characters associated with *míng* 明 and /-n/

| | | Persian | Uyghur | Korean | Total |
|---|---|---|---|---|---|
| ***míng* 明** | 密 | 11 | | | 11 |
| | 母 | 1 | | | 1 |
| | 木 | | 1 | | 1 |
| | 門 | | | 3 | 3 |
| **/-n/** | 音 | 12 | | | 12 |

According to the criteria proposed above, only *mì* 密 qualifies as an ST character in the *míngmŭ* 明母 group. However, *mù* 木 deserves special attention, since it was used in the *jīlín-lèishì* 鷄林類事 to represent the Korean coda /-m/ (Kang 1980). The ST character containing the *yùnwěi* 韻尾 /-n/ is *yīn* 音, a usage reminiscent of its function as a final consonant marker in *Hyangchal* 鄕札 transcription.

Table 37 below summarizes the distributional patterns of *míngmŭ* 明母 characters and transcription characters containing the *yùnwěi* 韻尾 /-n/ across the language sections.

**Table 37** Correspondence patterns of *míng* 明 and /-n/

| | Persian | Uyghur | Korean |
|---|---|---|---|
| ***míng* 明母** | m | m | m |
| **/-n/** | n | – | – |

The ST patterns associated with *míngmŭ* 明母 characters and transcription characters containing the *yùnwěi* 韻尾 /-n/ are relatively straightforward. As shown in Table 37, *míngmŭ* 明母 characters were used to represent m, whereas transcription characters containing the *yùnwěi* 韻尾 /-n/ were used to represent [n].

Table 38 presents a summary of the ST patterns identified in the present study.

## 5 Major characteristics of transcription in *Huìtóngguănxì Huáyíyìyŭ*

The analyses presented above reveal that the transcription system of HHY was governed by a set of systematic structural and phonetic principles recurring across the language sections. Although individual correspondences vary depending on the phonological properties of the target languages, the overall transcription system exhibits a high degree of internal consistency. In particular, two general characteristics emerge repeatedly throughout the corpus: phonetic transcription and consistent notation. These characteristics are not confined to individual language sections, but reflect broader principles underlying the HHY transcription system as a whole.

**Table 38** Summary of *shēngmŭ* 聲母 transcription patterns

| | | chun 脣 | | she 舌, chi 齒 | | | | ya 牙 | | hou 喉 |
|---|---|---|---|---|---|---|---|---|---|---|
| | | **Bilabial** | **Labio-dental** | **Dental** | **Alveolar** | **Post-alveolar** | **Alveo-palatal** | **Velar** | **Postvelar** | **Glottal** |
| **stop** | **Voiceless** | p 幇 | | | t 端透 | | | k 溪 | q 溪 | |
| | **Voiced** | b 幇 | | | d 端 | | | g 見 | | |
| **Affricate** | **Voiceless** | | | | | ʧ 穿 | | | | |
| | **Voiced** | | | | | ʤ 照 | | | | |
| **Fricative** | **Voiceless** | | f 非 | | s 心 | ʃ 審 | | x 曉 | | h 曉 |
| | **Voiced** | | | | z 精 | ʒ 日 | | ɣ 影見 | | |
| **Liquid** | **Trill** | | | | r 兒來 | | | | | |
| | **Tap** | | | | ɾ 兒來 | | | | | |
| | **Lateral approximant** | | | l 來兒 | | | | | | |
| **Nasal** | | m 明 | | | n /-n/ | | | | | |

## 5.1 Phonetic transcription

Phonetic transcription records perceived phonetic realizations without reducing them to abstract phonological categories. Unlike phonological transcription, which neutralizes contextual variation in order to represent underlying contrastive units, phonetic transcription aims to capture surface realizations more directly. English /t/ provides a familiar example. Although represented uniformly in orthography, its phonetic realization varies by context, appearing as [$t^h$] in *top*, [t] in *stop*, and [ɾ] in *butter*. A phonetic transcription system represents these differences explicitly.

The digitized HHY data suggest that the transcription system systematically recorded perceived phonetic realizations rather than abstract phonological categories. Rather than neutralizing contextual variation in order to represent underlying contrastive units, HHY transcription reflects surface realizations more directly. This sensitivity to phonetic detail is observed across multiple language sections and is consistent with previous descriptions of HHY as a resource oriented toward spoken language (Nishida 1963, p. 98; Ochi 2004, p. 115). Previous studies, however, focused primarily on individual language sections. The present study showed that this characteristic extends across the HHY corpus as a whole.

At the same time, the transcription characters used in HHY cannot be equated with the IPA. The IPA is a phonetic notation system in which each symbol corresponds to a single speech sound. The Chinese characters in HHY, by contrast, functioned primarily as a writing system for a natural language and therefore encoded broader phonological categories rather than discrete phonetic segments. As a result, when Chinese characters were used for phonetic transcription, phonetic variants that belong to a single phoneme in the target language could be represented separately, while sounds belonging to a single phonemes in the target language could conversely be grouped within the same *yīnxì* 音系 category.

Furthermore, the phonetic range associated with a given Chinese phonological category was not fixed, but shifted when that category was used to transcribe foreign languages. Such shifts emerged through interaction between the Chinese phonological system and the phonological systems of the target languages. This flexible use of Chinese phonological categories constitutes one of the defining characteristic of HHY transcription and is closely related to the purpose for which the corpus was compiled. As discussed above, HHY differs from the other three classes of *huáyíyìyǔ* 華夷譯語 in being explicitly oriented toward spoken language, reflecting its primary role as a resource for training interpreters rather than translators.

## 5.2 Consistent notation

Our analysis of the digitized HHY data also reveals a high degree of consistency in notation. Like the phonetic orientation of HHY transcription, this characteristic has previously been noted in studies of individual language sections (Nishida 1963, p. 99), and the present study confirms that it extends across the corpus as a whole. Identical morphemes tend to be transcribed in largely the same way even when they occur within larger constructions, such as compound words or phrases.

One notable exception to this general consistency involves the use of ST. While the MT characters remain stable, the presence or absence of ST may vary. When the lexical identity of an item is sufficiently clear without ST, ST characters tend to be omitted; otherwise, they tend to be retained. Table 39 presents examples from the Korean section. Only a subset of cases in which the same lexical item occurs multiple times is included.

**Table 39** Examples of repeated lexical items in the Korean section

| Index | 1st row | 2nd row | Index | 1st row | 2nd row |
|---|---|---|---|---|---|
| K-1, 13–21, 232 | 天 | 哈嫩 (二) | K-10, 51 | 雪 | 嫩 |
| K-2, 22–27 | 日 | 害 | K-11, 50 | 霧 | 按蓋 |
| K-3, 28–31 | 月 | 得二 | K-12, 52–53 | 露 | 以沁 |
| K-4, 32–35 | 星 | 別二 | K-14, 124 | 陰 | 黑立大 |
| K-5, 36–38 | 風 | 把論 | K-17, 77, 108 | 高 | 那大 |
| K-6, 39–44 | 雲 | 故論 | K-18, 86 | 邊 | 格自 |
| K-7, 49 | 雷 | 別刺 | K-20, 127 | 晚 | 展根 (格) 大 |
| K-8, 45–48, 462 | 雨 | 必 | | | |

This pattern is clearly illustrated by the item *hānèn* 哈嫩 together with the ST character *èr* 二. The full form *hānèn-èr* 哈嫩二 appears only when the item occurs as an independent entry, as in K-1. When the same form appears within compounds or phrases, as in K-13–21 and K-232, the ST character *èr* 二, representing the coda /l/, is omitted, and the form appears simply as *hānèn* 哈嫩.

By contrast, in the cases of *dé-èr* 得二 and *bié-èr* 別二, where the same ST character *èr* 二 likewise represents /l/, the character is consistently retained throughout the corpus. This difference can be explained by lexical recoverability. Even without the ST character, *hānèn* 哈嫩 can still be interpreted unambiguously as representing Korean /hanal/ 'sky'. By contrast, *dé* 得 and *bié* 別 alone do not provide sufficiently transparent representations of Korean /tal/ 'moon'and /pjəl/ 'star'without the additional ST character *èr* 二.

Although the number of cases is limited, a similar pattern is also observed with grammatical morphemes. Grammatical morphemes that are transcribed consistently across entries in Korean section are summarized in Table 40.

**Table 40** Grammatical morphemes consistently transcribed in the Korean section

| 2nd row | Korean | Index |
|---|---|---|
| 大 | *-ta* | K-13–14, 17, 19–35, 39–40, 47–55, 77–78, 84, 87–88, 91–100, 106–113, 118–119, 124, 127–129, 134–135, 142–143, 176, 188–189, 218–219, 229, 253–254, 264–265, 353–355, 357–358, 381–383, 387, 426–427, 436–437, 526–527 (88 items in total) |
| 剌 | *-la* | K-146–147, 190, 341, 342, 345, 347, 349–350, 356, 506, 507 (12 items in total) |
| 格 | *-ge~e* | K-20–23, 29–30, 50, 146–147, 189, 341, 345, 347, 349–350 (15 items in total) |
| 那 | *-na* | K-190, 342 (2 items in total) |

The character *dà* 大, for example, is used to transcribe two distinct sentence-final endings in Middle Korean: *-ta*, the declarative ending, and *-la*, the imperative ending. The forms *-ke-* and *-e-*, transcribed with *gé* 格, as well as *-na-*, transcribed with *nà* 那, function as confirmative prefinal endings and are typically associated with monologic speech or utterances with a strongly one-sided assertive function (Ko 2010, pp. 285–286).

This distinction is reflected in the transcription of *zhǎngēn(gé)dà* 展根 (格) 大 in the Korean section. In the form *zhǎngēndà* 展根大 'to grow dark', corresponding to Middle Korean *cyengkul-ta*, the ST character *gé* 格, representing the confirmative prefinal ending -*e-*, is omitted. By contrast, in *zhǎngēngédà* 展根格大 'to have grown dark', corresponding to Middle Korean *cyengkul-e-ta*, *gé* 格 is inserted to explicitly mark the presence of the prefinal ending.

# 6 General Principles Governing Main and Supplementary Transcription

The distribution of phonetic material between MT and ST in HHY is highly systematic across the language sections. In particular, the distribution of ST reveals systematic constraints on how phonetic information was encoded within the HHY transcription system. These constraints reflect both the structural limitations of the Chinese syllable template and the phonetic properties of the target languages.

## 6.1 General principle I: syllable-structure constraints on Main Transcription

Differences between the phonological constraints of Chinese and those of the target languages played a crucial role in shaping HHY transcription practices. The first general principle governing HHY transcription is grounded in syllable-structure constraints and is best understood in relation to the Chinese syllable template, conventionally formalized as I/MVET (Initial = *shēngmǔ* 聲母; Medial = *yùntóu* 韻頭; principal Vowel = *yùnfù* 韻腹; Ending = *yùnwěi* 韻尾; Tone = *shēngdiào* 聲調).

Within this framework, the *yùntóu* 韻頭 and *yùnfù* 韻腹 do not play a major role in determining the distribution of MT. Instead, constraints on MT are associated primarily with the *shēngmǔ* 聲母 and the *yùnwěi* 韻尾. This asymmetry corresponds closely to specific properties of the Chinese syllable structure in the fifteenth and sixteenth centuries, two of which are particularly relevant here. **(a)** Fifteenth- and sixteenth-century Chinese permitted only /n/ and /ŋ/ as codas. **(b)** Complex initials were not permitted. Together, these structural constraints placed clear limits on the range of phonetic material that could be represented by a single Chinese character in HHY.

Under constraints (a) and (b), a range of target-language phonological configurations could not be represented by MT. First, MT was systematically limited in word-final position due to constraint (a). The data show that MT was not possible when a target-language syllable or word ended in a consonant other than /n/ or /ŋ/. Consonants such as /k/, /t/, /p/, /h/, /l/, and /r/ therefore consistently fell outside the range of coda representation by MT characters. However, the *yùnwěi* 韻尾 /m/, generally assumed to have disappeared from Chinese by the fifteenth century, is occasionally represented by MT in a small number of HHY entries. It remains unclear whether this reflects residual survival of /-m/ in Chinese or localized transcriptional variation.

MT was likewise systematically limited in word-initial position due to constraint (b). Because complex onsets were excluded from the Chinese syllable template, word-initial consonant clusters in the target languages could not be fully encoded by MT. In such cases, the first consonant in the cluster was consistently excluded from MT and instead recovered through ST, while the second consonant could be represented as an initial.

MT was also limited in word-medial position due to both constraints (a) and (b). When three or more consonants occurred medially, only a subset could be encoded by MT. If the first consonant was /n/ or /ŋ/, it could be represented as a coda, while any intervening consonants were excluded from MT; the final consonant, however, could still be represented as an initial. Word-final consonant clusters exhibited comparable limitations. Only the first consonant could be encoded as a coda, and only when it was /n/ or /ŋ/, while all remaining consonants could not be represented by MT.

Taken together, these patterns show that sounds that could not be encoded by MT under the constraints of the Chinese syllable structure systematically satisfied the first necessary condition for ST. The structural constraints identified here can be summarized schematically as follows.

$$\mathbf{C}_1\ \mathbf{C}_2\ \mathbf{V}\ \mathbf{C}_3\ \mathbf{C}_4\ \mathbf{C}_5\ \mathbf{C}_6\ \mathbf{V}\ \mathbf{C}_7\ \mathbf{C}_8$$

When the constraints outlined above are applied to this schematic representation, $C_1$, $C_4$, $C_5$, and $C_8$ consistently qualify as potential targets for ST regardless of their specific phonetic value. By contrast, $C_2$ and $C_6$ are eligible for MT. $C_3$ and $C_7$ are encoded as MT only when they are [n] or [ŋ]; otherwise, they likewise satisfy the first necessary condition for ST.

Crucially, however, not all such sounds were realized through ST. Additional phonetic conditions also had to be met for ST to occur, as discussed in the following section.

## 6.2 General principle II: Phonetic conditions governing Supplementary Transcription

While General Principle I specifies which phonetic material cannot be encoded by MT under the constraints of the Chinese syllable structure, the HHY data further show that only a subset of such sound was actually realized through ST. The distribution of ST was therefore not arbitrary, but correlated systematically with specific phonetic properties of the target-language segments.

Across language sections, segments realized through ST consistently exhibited one or more of the following phonetic characteristics. First, voiced segments are regularly represented by ST whenever they satisfied the structural conditions identified in General Principle I. No voiced segment attested in the phoneme inventories of the target languages was systematically excluded from ST under these conditions. Apparent absences reflect gaps in the relevant inventories rather than restrictions on ST itself. Previous studies have used the presence of ST in HHY as independent evidence for the voicing of particular segments (e.g. Shōgaito 1984; Ochi 2004).

Second, voiceless segments may also be realized through ST when they are continuants. Under the same structural constraints, voiceless fricatives and other continuant segments consistently appeared as ST where MT was unavailable. As with voiced segments, no systematic exclusion of voiceless continuants from ST is observed in the HHY data.

Third, even voiceless stops may be realized through ST when they are phonetically released. Evidence from *Hunminjeongeum Haeryebon* 訓民正音解例本, particularly the Eight Coda Rules (八終聲法), suggests that fifteenth-century Korean permitted distinctions involving unreduced or unreleased codas (Lee and Ramsey 2011). Persian, Uyghur, and Cham sections also provide clear cases in which released voiceless stops are represented by ST under positions where MT is structurally unavailable.

Taken together, these observations indicate that, provided the syllable-structure constraints outlined in General Principle I were satisfied, segments were eligible for ST if they met at least one of the following conditions:

**A. The segment is voiced.**
**B. The segment is voiceless but continuant.**
**C. The segment is a voiceless stop realized with release.**

Although a small number of cases do not conform neatly to these generalizations, such instances are sporadic and do not undermine the overall regularity observed in the HHY transcription system. The patterns summarized here therefore capture the dominant phonetic conditions governing the distribution of ST across the corpus.

## 7 General Discussion and Conclusion

This study has argued that HHY should be understood not as a collection of independent language glossaries, but as a coherent multilingual transcription system. Through a cross-linguistic comparison of multiple language sections, the analysis has shown that Chinese characters were used systematically to represent foreign-language speech. The regularities identified in the distribution of *shēngmŭ* 聲母 and *yùnmŭ* 韻母 categories, as well as in the use of MT and ST, demonstrate that HHY was governed by stable transcriptional principles rather than by accidental or purely orthographic variation.

A central result of the study is that the Chinese phonological categories used in HHY did not function as rigid reflections of native Chinese phonology. Previous scholarship has often interpreted HHY transcriptions directly through reconstructed Chinese phonological systems, especially through Late *Míng* 明 *yīnxì* 音系 categories reconstructed from rime books and related sources. Such an approach implicitly assumes that the phonetic values of Chinese characters remained stable when they were used to transcribe foreign languages. The evidence examined here suggests otherwise. In HHY, Chinese phonological categories functioned as flexible phonetic approximation devices whose ranges shifted according to the phonological structure of the target language.

This finding has methodological implications for the study of premodern multilingual transcription materials. HHY records spoken forms from multiple languages within a shared transcriptional framework, and this makes it possible to analyze the transcription system itself. Correspondences that remain ambiguous within a single language section often become interpretable when examined against cross-linguistic regularities. Languages with relatively secure historical phonologies, such as Korean and Japanese, can therefore serve as calibration points for interpreting less well-documented sections. In this respect, HHY offers a type of evidence that is unavailable in monolingual transcription materials.

The comparative framework proposed here also changes how language-specific HHY studies can be used. Previous research has largely remained confined to individual language traditions, with results rarely integrated across fields. By treating HHY as a unified transcriptional system, findings from better-documented languages can be used to inform the interpretation of less well-documented languages through shared transcriptional correspondences. HHY thus provides a comparative bridge through which language-specific reconstructions can acquire broader implications for the historical phonology of Asian languages.

The analysis also contributes to historical Chinese phonology. The practical phonetic ranges associated with Late *Míng* 明 transcription categories appear to have been broader and more flexible than native Chinese phonological descriptions alone would suggest. The use of *shēngmǔ* 聲母 and *yùnmǔ* 韻母 categories in HHY was shaped not only by Chinese syllable structure, but also by cross-linguistic perceptual approximation. HHY therefore provides indirect evidence for how *Míng* 明 interpreters perceived, categorized, and represented foreign sounds using the phonological resources available to them.

Several limitations also remain. The reliability of previous reconstructions differs substantially across the language sections, and some sections remain much less securely understood than others. Multiple manuscript traditions also introduce textual uncertainty in certain cases, while some transcription patterns, especially those involving ST, cannot yet be interpreted with complete confidence. The reconstruction proposed here should therefore be understood as probabilistic rather than definitive. Even so, the recurrence of comparable patterns across independent language sections strongly suggests that the major transcriptional correspondences identified in this study reflect genuine structural properties of the HHY system.

Overall, this study shows that HHY is an important but still underused resource for Asian historical phonology. Because it brings together languages with different degrees of historical documentation, including poorly attested or extinct languages, it provides a rare record of how foreign speech was perceived and represented across languages in the fifteenth and sixteenth centuries. As further reconstructions and diachronic studies become available, HHY can serve not only as a source for individual language histories, but also as a comparative foundation for studying premodern Asian phonology across language boundaries.